\documentclass[11pt,a4paper]{article}
\usepackage[hyperref]{acl2019}
\usepackage{times}
\usepackage{latexsym}
\usepackage{dialogue}
\usepackage{url}

\usepackage{graphicx}
\usepackage{enumitem}
\usepackage{booktabs}
\usepackage{color}
\usepackage{subfig}

\usepackage[draft,textsize=footnotesize]{todonotes}

\aclfinalcopy 

\title{The PhotoBook Dataset:\\ Building Common Ground through Visually-Grounded Dialogue}

\author{Janosch Haber$^*$ Tim Baumg\"artner$^\clubsuit$ Ece Takmaz$^\clubsuit$ Lieke Gelderloos$^\ddagger$\\ {\bf Elia Bruni}$^\dagger$ \and {\bf Raquel Fern\'andez$^\clubsuit$}\\
$^\clubsuit$University of Amsterdam, $^*$Queen Mary University of London\\ 
$^\ddagger$Tilburg University, $^\dagger$Universitat Pompeu Fabra\\
\texttt{\{raquel.fernandez|ece.takmaz\}@uva.nl}\\
\texttt{j.haber@qmul.ac.uk}, 
 \texttt{\{baumgaertner.t|elia.bruni\}@gmail.com}\\
 \texttt{l.j.gelderloos@tilburguniversity.edu}\\
 }

\date{}

\begin{document}
\maketitle

\begin{abstract}
This paper introduces the PhotoBook dataset, 
a large-scale collection of visually-grounded, task-oriented dialogues in English designed to investigate shared dialogue history 
accumulating during conversation.
Taking inspiration from seminal work on dialogue analysis,
we propose a data-collection task formulated as a collaborative game 
prompting two online participants to refer to images utilising both 
their visual context as well as previously established referring expressions. 
We provide a detailed description of the task setup and 
a thorough analysis of the 2,500 dialogues collected. 
To further illustrate the novel features of the dataset, 
we propose a baseline model for reference resolution which uses
a simple method to take into account shared information accumulated in a reference chain. 
Our results show that this information is particularly important to resolve later descriptions
and underline the need to develop more sophisticated models of common ground in dialogue interaction.\footnote{The PhotoBook dataset is being released by the Dialogue Modelling Group led by Raquel Fern\'andez at the University of Amsterdam. The core of this work was done while Janosch Haber and Elia Bruni were affiliated with the group.} 
\end{abstract}


\section{Introduction}
\label{sec:intro}

The past few years have seen an increasing interest in developing 
computational agents for \textit{visually grounded dialogue}, the task of using natural language to communicate about visual content in a multi-agent setup. 
The models developed for this task often focus on specific aspects such as
image labelling~\cite{Mao2015, Vedantam2017}, object reference \cite{Kazemzadeh2014ReferIt, deVries2016}, visual question answering~\cite{Antol2015},
and first attempts of visual dialogue proper \cite{Das2016},
 but fail to produce consistent outputs over a 
 conversation. 

We hypothesise that one of the main reasons for this shortcoming is  
the models' inability to effectively utilise dialogue history. Human interlocutors are known 
to collaboratively establish a shared repository of mutual information during a conversation \cite{Clark1986, Clark1996, Brennan1996}. This \textit{common ground} 
\cite{Stalnaker1978} then is used to optimise understanding and communication efficiency. 
Equipping artificial dialogue agents with a similar representation of dialogue context
thus is a pivotal next step in improving the quality of their dialogue output.

\begin{figure*}[t]\centering
    \includegraphics[width=\linewidth]{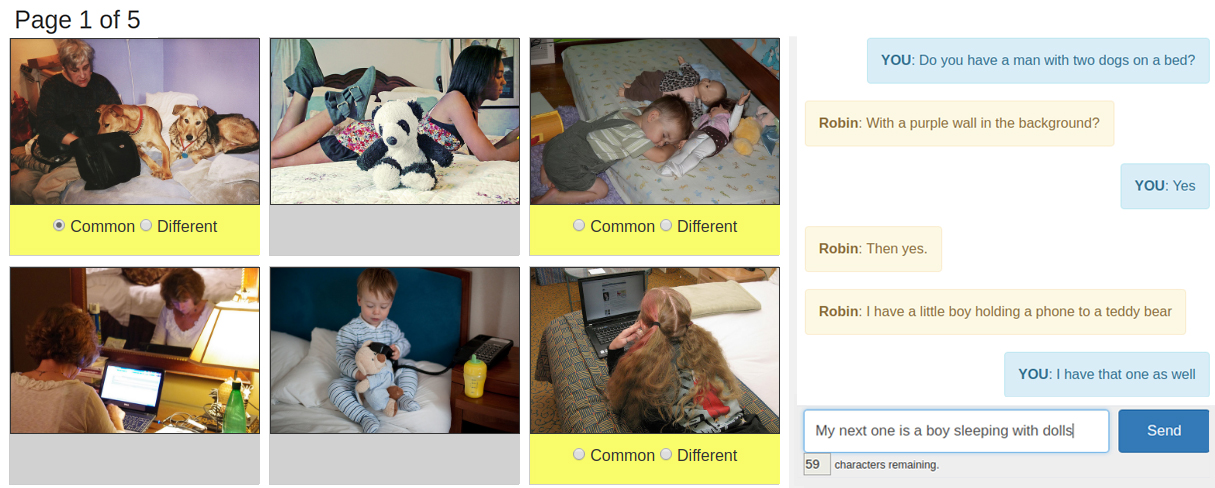} 
    \caption{Screenshot of the Amazon Mechanical Turk user interface designed to collect the PhotoBook dataset.
        }
    \label{fig:screenshot}
\end{figure*}

To facilitate progress towards more consistent and effective conversation models, 
we introduce the \textit{PhotoBook} 
dataset:~a large collection of 2,500 human-human goal-oriented English conversations between two participants,
who are asked to identify shared images  in their respective photo books by 
exchanging messages via written chat. 
This setup takes inspiration from experimental paradigms extensively used within the psycholinguistics literature 
to investigate partner-specific common ground \cite[for an overview, see][]{brown2015people}, 
adapting them to the requirements imposed by online crowdsourcing methods.  
The task is formulated as a game consisting of five rounds. 
Figure~\ref{fig:screenshot} shows an example of a participant's display.
Over the five rounds of a game, a selection of previously displayed images will be visible again, 
prompting participants to re-refer to images utilising both their visual context as well as previously established referring expressions. 
The resulting dialogue data therefore allows for tracking the common ground developing 
between dialogue participants.

We describe in detail the PhotoBook task and the data collection, 
and present a thorough analysis of the dialogues in the dataset. In addition, to showcase how the new dataset may be exploited for computational modelling, 
we propose a reference resolution baseline model trained to identify target images 
being discussed in a given dialogue segment. The model
 uses a simple method to take into account information accumulated in a reference chain. 
Our results show that this information is particularly important to resolve later descriptions
and highlight the importance of developing more sophisticated models of common ground in dialogue interaction.

The PhotoBook dataset, together with the data collection protocol, the automatically extracted reference chains, and the code used for our analyses and models 
are available at the following site: \url{https://dmg-photobook.github.io}.


\section{Related Work}
\label{sec:related}

Seminal works on cooperative aspects of dialogue  
have developed their hypotheses and models based on a relatively small number of samples collected through lab-based conversation tasks \cite[e.g.,][]{Krauss1964, Krauss1966, Clark1986, Brennan1996,anderson1991hcrc}. 
Recent datasets inspired by this line of work include the REX corpora \cite{takenobu2012rex} and PentoRef \cite{Zarriess2016}. 
With the development of online data collection methods \cite{vonAhn2006} a new, game-based approach to quick and inexpensive collection of dialogue data became available.
PhotoBook builds on 
these traditions to provide a large-scale dataset suitable for data-driven development of computational dialogue agents.

The computer vision community has recently developed large-scale datasets for visually grounded dialogue \cite{Das2016,deVries2017}. These approaches extend earlier work on visual question answering \cite{Antol2015} to a multi-turn setup where two agents, each with a pre-determined Questioner or Answerer role, exchange sequences of questions and answers about an image. While data resulting from these tasks provides interesting opportunities to investigate visual grounding, it suffers from fundamental shortcomings with respect to the collaborative aspects of natural goal-oriented dialogue (e.g., fixed, pairwise structuring of question and answers, no extended dialogue history).
In contrast, PhotoBook includes natural and free-from dialogue data with a variety of dialogue acts and opportunities for participant collaboration.

Resolving referring expressions in the visual modality has also been studied in computer vision. Datasets such as ReferIt \cite{Kazemzadeh2014ReferIt}, Flicker30k Entities \cite{plummer2015flickr30k} and Visual Genome \cite{VisualGenome} map referring expressions to regions in a single image. 
Referring expressions in the PhotoBook dataset differ from this type of data in that the candidate referents are independent but similar images 
and, most importantly, are often part of a reference chain in the participants' dialogue history.

\section{Task Description and Setup}
\label{sec:task}

In the PhotoBook task, two participants are paired for an online multi-round image identification game. In this game, participants are shown collections of images that resemble the page of a photo book (see Figure~\ref{fig:screenshot}). Each of these collections is a randomly ordered grid of six similar images depicting everyday scenes extracted from the MS COCO Dataset \cite{LinCOCO2014}. On each page of the photo book, some of the images are present in the displays of both participants (the \textit{common} images). The other images are each shown to one of the participants only (\textit{different}). Three of the images in each display are highlighted through a yellow bar under the picture. The participants are tasked to mark these highlighted  target images as either \textit{common} or \textit{different} by chatting with their partner.\footnote{Pilot studies showed that labelling all six images took participants about half an hour, which appeared to be too long for the online setting, resulting in large numbers of disconnects and incomplete games.}
The PhotoBook task is symmetric, i.e., participants do not have predefined roles such as instruction giver and follower, or questioner and answerer. 
Consequently, 
both participants can freely and naturally contribute to the conversation, leading to more natural dialogues. 

Once the two participants have made their selections on a given page, they are shown a feedback screen and 
continue to the next round of the game, a new page of the photo book displaying a different grid of images. Some of the images in this grid  will be new to the game while others will have appeared before. 
A full game consists of labelling three highlighted target images in each of five consecutive rounds.

Each highlighted image is displayed exactly five times throughout a game while the display of images and the order of rounds is randomised to prevent participants from detecting any patterns. 
As a result of this carefully designed setup, dialogues in the PhotoBook dataset 
contain multiple descriptions 
of each of the target images and thus provide a valuable resource for investigating participant cooperation, and specifically collaborative referring expression generation and resolution with respect to the conversation's common ground.

\paragraph{Image Sets}
The task setup requires each game of five rounds to display 12 unique but similar images to elicit non-trivial referring expressions. 
We use the object category annotations in MS COCO to group all landscape, unmarked, colour images where the two largest objects belong to the same category across all images in the set (e.g., all images in the set prominently feature a person and a cat).\footnote{All images where the two largest objects cover less than 30k pixels ($\sim$10\% of an average COCO image) were rejected.}
This produced 30 sets of at least 20 images 
 from which 12  were selected at random. As a given game highlights only half of the images from a given set, each image set produces two different game sets with different target images to be highlighted, for a total of 60 unique games and 360 unique images.
 More details on the PhotoBook setup and image sets are provided in Appendix~\ref{app:task}.

\section{Data Collection}
\label{sec:collection}

We use the ParlAI framework~\cite{Miller2017} to implement the task and interface with crowdsourcing platform
Amazon Mechanical Turk (AMT) to collect the data. To control the quality of collected dialogues, we require AMT workers to be native English speakers and to have completed at least 100 other tasks on AMT with a minimum acceptance rate of 90\%. Workers are paired through an algorithm based on whether or not they have completed the PhotoBook task before and which of the individual games they have played. In order to prevent biased data, workers can complete a maximum of five games, each participant can complete a given game only once, and the same pair of participants cannot complete more than one game.

Participants are instructed about the task and first complete a warming-up round with only three images per participant  (two of them highlighted). In order to render reference grounding as clean as possible and facilitate automatic processing of the resulting dialogue data, participants are asked to try to identify the common and different images as quickly as possible, only describe a single image per message, and directly select an image's label when they agree on it.
The compensation scheme is based on an average wage of 10 USD per hour \cite{Hara2017}.
See Appendix~\ref{app:collection} for a full account of the instructions and further details on participant payment.

\begin{figure*}[t]
     \subfloat[\label{subfig-4:efficiency}]{%
       \includegraphics[height=4.3cm]{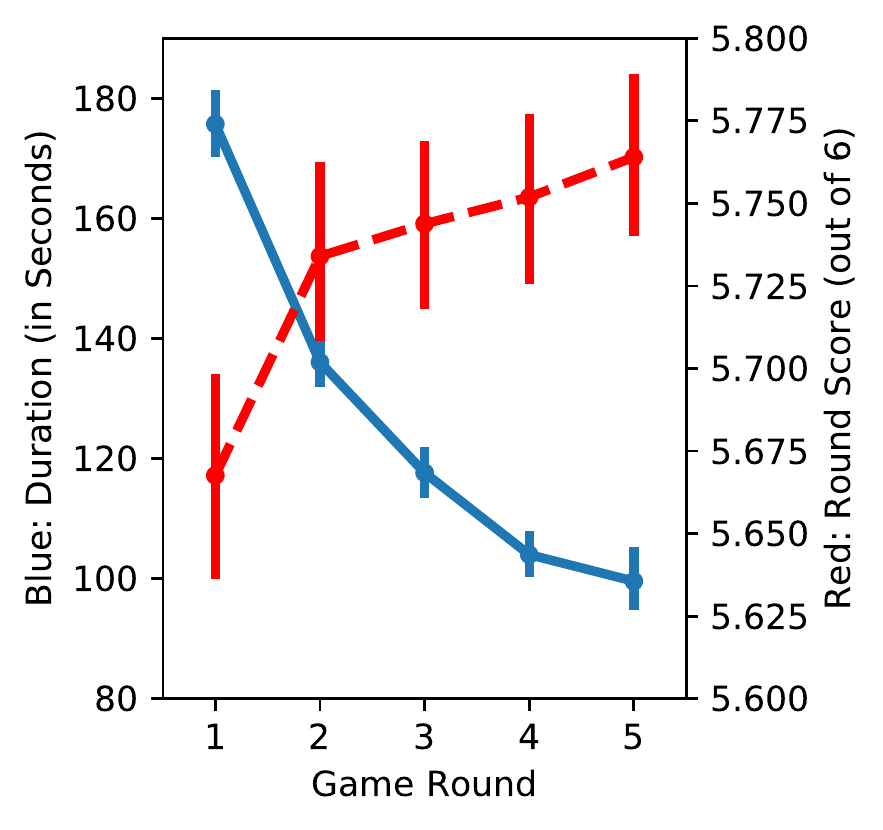}
     }
     \hfill
    \subfloat[\label{subfig-2:ratio}]{%
       \includegraphics[height=4.3cm]{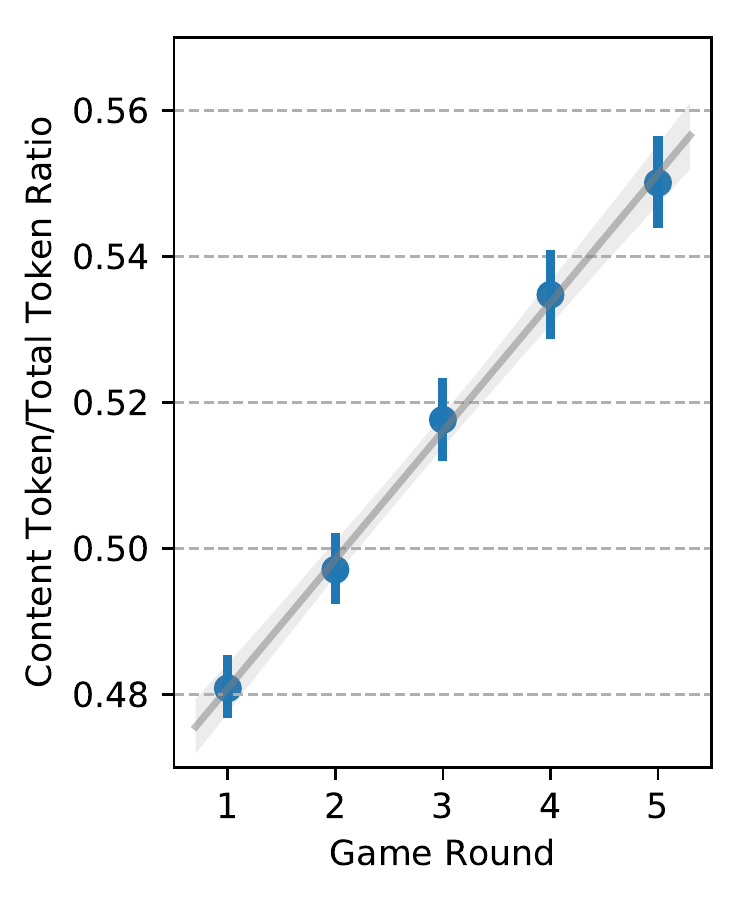}
     }
     \hfill
     \subfloat[\label{subfig-1:new_tokens}]{%
       \includegraphics[height=4.3cm]{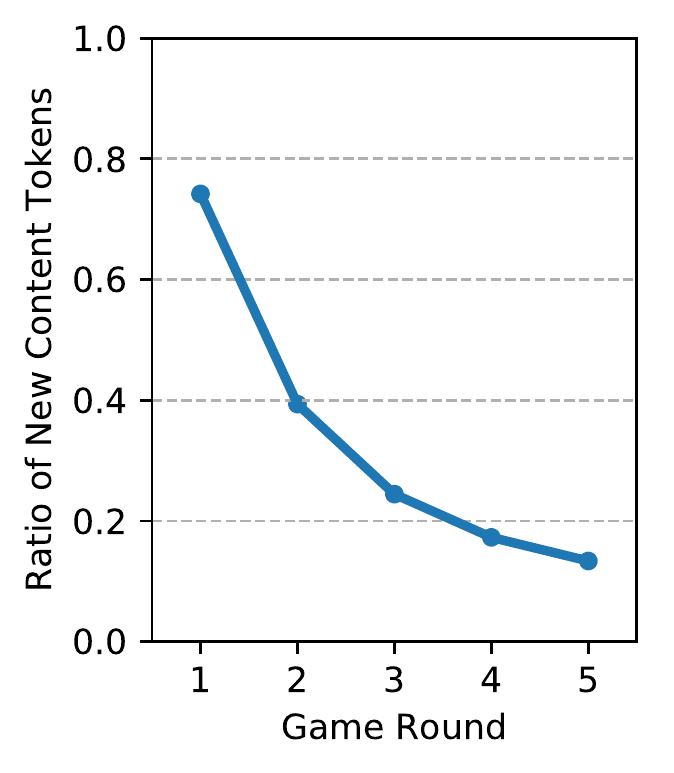}
     }
     \hfill
     \subfloat[\label{subfig-3:pos_tag_change}]{%
       \includegraphics[height=4.3cm]{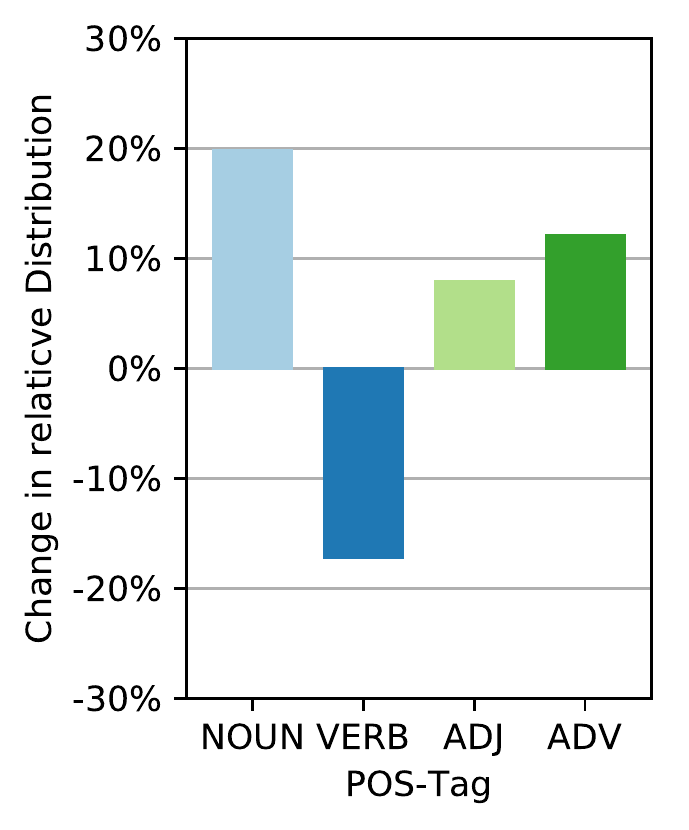}
     }
     \vspace*{-5pt}
     \caption{
     	{\bf (a)}~Average completion times (solid blue) and scores  (dashed red) per game round. 
     	{\bf (b)}~Ratio of content tokens over total token count per round with best linear fit.
	{\bf (c)}~Ratio of new content tokens over total content token count per round.
	{\bf (d)}~Relative change in distribution of main content POS between the first and last game round.}
     \label{fig:feedback}
\end{figure*}

During data collection, we recorded anonymised participant IDs, 
the author, timestamp and content of all sent messages, label selections and button clicks, 
plus self-reported collaboration performance scores. 
For a period of two months, data collection produced human-human dialogues for a total of 2,506 completed games. 
The resulting PhotoBook dataset contains a total of 164,615 utterances, 130,322 actions and spans a vocabulary of 11,805 unique tokens. 

Each of the 60 unique game sets was played between 15 and 72 times, with an average of 41 games per set.
The task was completed by 1,514 unique workers, of which 472 only completed a single game, 448 completed between two and four games, and 594 the maximum of five games. Completing a full five-round game took an average of 14.2 minutes. 
With three highlighted images per player per round, during a full game of five rounds 30 labelling decisions have to be made. On average, participants correctly labelled 28.62 out of these 30.


\section{Dataset Analysis}
\label{sec:data}

In this paper, we focus on the analysis of participants' interaction during a game of five labelling rounds.\footnote{Tracking participant IDs, for example, also allows for an analysis of differences in behaviour across different games.} Our data here largely confirms the observations concerning participants' task efficiency and language use during a multi-round communication task made by seminal, small-scale experiments such as those by \citet{Krauss1964, Clark1986, Brennan1996} and, 
due to its scale, offers additional aspects for further investigation.

\subsection{Task Efficiency}
Completing the first round of the PhotoBook task takes participants an average of almost three minutes. Completing the fifth round on the other hand takes them about half that time. As Figure \ref{subfig-4:efficiency} shows, this decline roughly follows a negative logarithmic function, with significant differences between rounds 1, 2, 3 and 4, and plateauing towards the last round. The number of messages sent by participants as well as the average message length follow a similar pattern, significantly decreasing between consecutive game rounds. 
The average number of correct image labels, on the other hand, significantly increases between the first and last round of the game (cf.~the red dashed graph in Figure \ref{subfig-4:efficiency}).
As a result, task efficiency as calculated by points per minute significantly increases with each game round.

\subsection{Linguistic Properties of Utterances}
To get a better understanding of how participants increase task efficiency and shorten their utterances, we analyse how the linguistic characteristics of messages change over a game. 

We calculated a normalised content word ratio by dividing the count of content words by the total token count.\footnote{We filtered out function words with NLTK's stopword list \url{http://www.nltk.org/}.} 
This results in an almost linear increase of content tokens over total token ratio throughout a game (average Pearson's $r$ per game of 0.34, $p\!\!\ll\!\!0.05$, see Figure \ref{subfig-2:ratio}). With referring expressions and messages in general getting shorter, content words thus appear to be favoured to remain.
We also observe that participants reuse these content words. 
Figure \ref{subfig-1:new_tokens} shows the number of novel content tokens per game round, which 
roughly follows a negative logarithmic function. This supports the hypothesis of participants establishing a 
\textit{conceptual pact} on the referring expression attached to a specific referent: Once accepted, a referring expression is typically refined through shortening rather than by reformulating or adding novel information  \cite[cf.,][]{Brennan1996}. 

We also analysed in more detail the distribution of word classes per game round by tagging messages with the NLTK POS-Tagger.
Figure \ref{subfig-3:pos_tag_change} displays the relative changes in content-word-class usage between the first round and last round of a game. 
All content word classes but verbs show a relative increase in occurrence, most prominently 
nouns  with a 20\% relative increase. The case of adverbs, which show a 12\% relative increase, is particular: Manual examination showed that most adverbs are not used to described images but rather to flag that a given image has already appeared before or to confirm/reject (`again' and `too' make up 21\% of all adverb occurrences; about 36\% are `not', `n't' and `yes'). 
These results indicate that interlocutors are most likely to retain the nouns and adjectives of a developing referring expression, while increasingly dropping verbs, as well as prepositions and determiners. A special role here takes definite determiner `the', which, in spite of the stark decline of determiners in general, increases by 13\% in absolute occurrence counts between the first and last round of a game, suggesting a shift towards known information.

Finally, in contrast to current visual dialogue datasets \cite{Das2016,deVries2017} which exclusively contain sequences of question-answer pairs, the 
PhotoBook dataset includes diverse dialogue acts.
Qualitative examination shows that,  not surprisingly, 
a large proportion of messages include an image description. These descriptions however are interleaved with clarification questions, acceptances/rejections, and acknowledgements. For an example, see the dialogue excerpt in Figure \ref{fig:screenshot}. 
Further data samples are available in Appendix~\ref{app:data}.
A deeper analysis of the task-specific dialogue acts would require manual annotation, which could be added in the future.

\subsection{Reference Chains}
\label{sec:data-chains}

In a small-scale pilot study, 
\citet{ilinykh2018task} find that the pragmatics of goal-oriented dialogue leads to consistently more factual scene descriptions and reasonable referring expressions than traditional, context-free labelling of the same images. 
We argue that in the PhotoBook task referring expressions are not only adapted based on the goal-oriented nature of the interaction but also by incorporating the developing common ground between the participants. This effect becomes most apparent when collecting all referring expressions for a specific target image produced during the different rounds of a game in its
\textit{coreference chain}.
The following excerpt displays such a coreference chain extracted from the PhotoBook dataset:
\begin{enumerate}[itemsep=-1pt,leftmargin=14pt]
\item \textbf{A:} \textit{Do you have a boy with a teal coloured shirt with yellow holding a bear with a red shirt?} 
\item \textbf{B: }\textit{Boy with teal shirt and bear with red shirt?}  
\item \textbf{A: }\textit{Teal shirt boy?} 
\end{enumerate}
\noindent To quantify the 
effect of referring expression refinement, 
we compare participants' first and last descriptions to a given target image with the image's captions provided in the MS COCO dataset. For this purpose we manually annotated the first and last expressions referring to a set of six target images across ten random games in the PhotoBook dataset. 
Several examples are provided in Appendix~\ref{app:data}. 
Table \ref{tab:coco_comparison} shows their differences in token count before and after filtering for content words with the NLTK stopword list.

\begin{table}[ht]\centering
\scalebox{0.88}{\begin{tabular}{|l|c|c|c|c|}
\hline
\textbf{Source}   & \multicolumn{1}{l|}{\textbf{\# Tokens}} & \textbf{\# Content} & \textbf{Distance} \\ \hline
COCO captions   & 11.167      & 5.255      &   --  \\ \hline
First description & 9.963       & 5.185    &   0.091   \\ \hline
Last description  & 5.685       & 5.128   &   0.156  \\ \hline         
\end{tabular}}
\caption{Avg.~token counts in COCO captions and the first and last descriptions in PhotoBook, plus their cosine distance to the caption's cluster mean vector. The distance between first and last descriptions is 0.083.}
\label{tab:coco_comparison}
\end{table}

Before filtering, first referring expressions do not significantly differ in length from the COCO captions.
Last descriptions however are significantly shorter than both the COCO captions and first descriptions. After filtering for content words, no significant differences remain. 
We also calculate the cosine distance between the three different descriptions based on their average word vectors.\footnote{We average pretrained word vectors from Word2Vec \cite{Mikolov2013} in gensim (\url{https://radimrehurek.com/gensim/}) to generate utterance vectors. The five COCO captions are represented by a cluster mean vector.}
Non-function words here should not significantly alter an utterance's mean word vector, which is confirmed in our results.
Before as well as after filtering, the distance between last referring expression and COCO captions is almost double the distance between the first referring expressions and the captions (see last column in Table~\ref{tab:coco_comparison}).
Comparing  the distribution of word classes in the captions and referring expressions finally revealed a similar distribution in first referring expressions and COCO captions, and a significantly different distribution in last referring expressions, among other things doubling the relative frequency of nouns.


\section{Reference Chain Extraction}
\label{sec:chains}

Collecting reference chains from dialogue data is a non-trivial task which normally requires manual annotation \cite{yoshida2011referring}. 
Here we propose a simple procedure to automatically extract reference chains made up of dialogue segments.  
A dialogue segment is defined as a collection of consecutive utterances that, as a whole, 
discuss a given target image and include expressions referring to it.
All dialogue segments within a game that refer to the same target image form its reference chain.

In order to automatically segment the collected dialogues in this way, we developed a rule-based heuristics exploiting participants' image labelling actions to detect segment boundaries and their respective targets.
The heuristics is described in detail in Appendix~\ref{app:chains}.
Since the task instructs participants to label images as soon as they identify them as either common or different, the majority of registered labelling actions can be assumed to conclude the current dialogue segment. 
The following excerpt displays a segment extracted from a game's first round, discussing one target image before a participant selects its label:

\begin{dialogue}
\speak{\textbf{B}} \textit{I have two kids (boys) holding surf boards walking.}
\speak{\textbf{A}} \textit{I do not have that one.}
\speak{\textbf{B}} \texttt{\small marks \#340331 as different}
\end{dialogue}

\noindent
Image selections however do not always delimit segments in the cleanest way possible. 
For example, a segment may refer to more than one target image, i.e., the participants may discuss two images and only after this discussion be able to identify them as common/different. 
72\% of the extracted segments are linked to only one target; 25\% to two.
Moreover, reference chains do not necessarily contain one segment for each of the five game rounds. 
They may contain fewer or more segments than rounds in a game, since participants may discuss
the same image more than once in a single round and some of the extracted 
chains may be noisy, as explained in the evaluation section below.
75\% of the automatically extracted chains contain three to six segments.

\paragraph{Evaluation} 
To evaluate the segmentation, two annotators independently reviewed segments extracted from 20 dialogues. These segments were annotated by marking all utterances $u$ in a segment $S$ with target images $I$ that refer to an image $i'$ where $i' \notin I$ to determine precision, and marking all directly preceding and succeeding utterances $u'$ outside of a segment $S$ that refer to a target image $i \in I$ to determine recall. Additionally, if a segment $S$ did not include any references to any of its target images $I$, it was labelled as improper.   
95\% of annotated segments were assessed to be proper (Cohen's $\kappa$ of 0.87), with 28.4\% of segments containing non-target references besides target references (Cohen's $\kappa$ of 0.97). Recall across all reviewed segments is 99\% (Cohen's $\kappa$ of 0.93).


\section{Experiments on Reference Resolution}
\label{sec:exp}

Using the automatically extracted dialogue segments, 
we develop 
a reference resolution model that aims at identifying the target images referred to in a dialogue segment. 
We hypothesise that later segments within a reference chain might be more difficult to resolve, because they rely on referring expressions previously established by the dialogue participants. As a consequence, a model that is able to keep track of the 
common ground should be less affected by this effect. To investigate these issues, we experiment with two conditions: 
In the {\sc No-History} condition, the model only has access to the current segment and to the visual features of each of the candidate images. 
In the {\sc History} condition, on the other hand, the model also has access to the previous segments in the
reference chain associated with each of the candidate images, containing 
the linguistic common ground built up by the participants. 

We keep our models very simple. Our aim is to propose baselines against which future work can be compared.

\subsection{Data}
\label{sec:exp-data}

The automatically extracted co-reference chains per target image were split into three disjoint sets for training (70\%), validation (15\%) and testing (15\%), aiming at an equal distribution of target image domains in all three sets. The raw numbers per data split are shown in Table~\ref{tab:splits}.

\begin{table}[b]\centering
\scalebox{0.86}{\begin{tabular}{|l|c|c|c|c|}
\hline
\bf Split & \bf Chains & \bf Segments & \bf Targets & \bf Non-Targets \\\hline
Train & 12,694 & 30,992 & 40,898 & 226,993 \\\hline
Val & 2,811 & 6,801 & 9,070 & 50,383 \\\hline
Test & 2,816 & 6,876 & 9,025 & 49,774 \\\hline
\end{tabular}}
\caption{Number of reference chains, dialogue segments, and image types (targets and non-targets)  in each data split.}
\label{tab:splits}
\end{table}

\begin{figure*}\centering
\includegraphics[width=.81\textwidth]{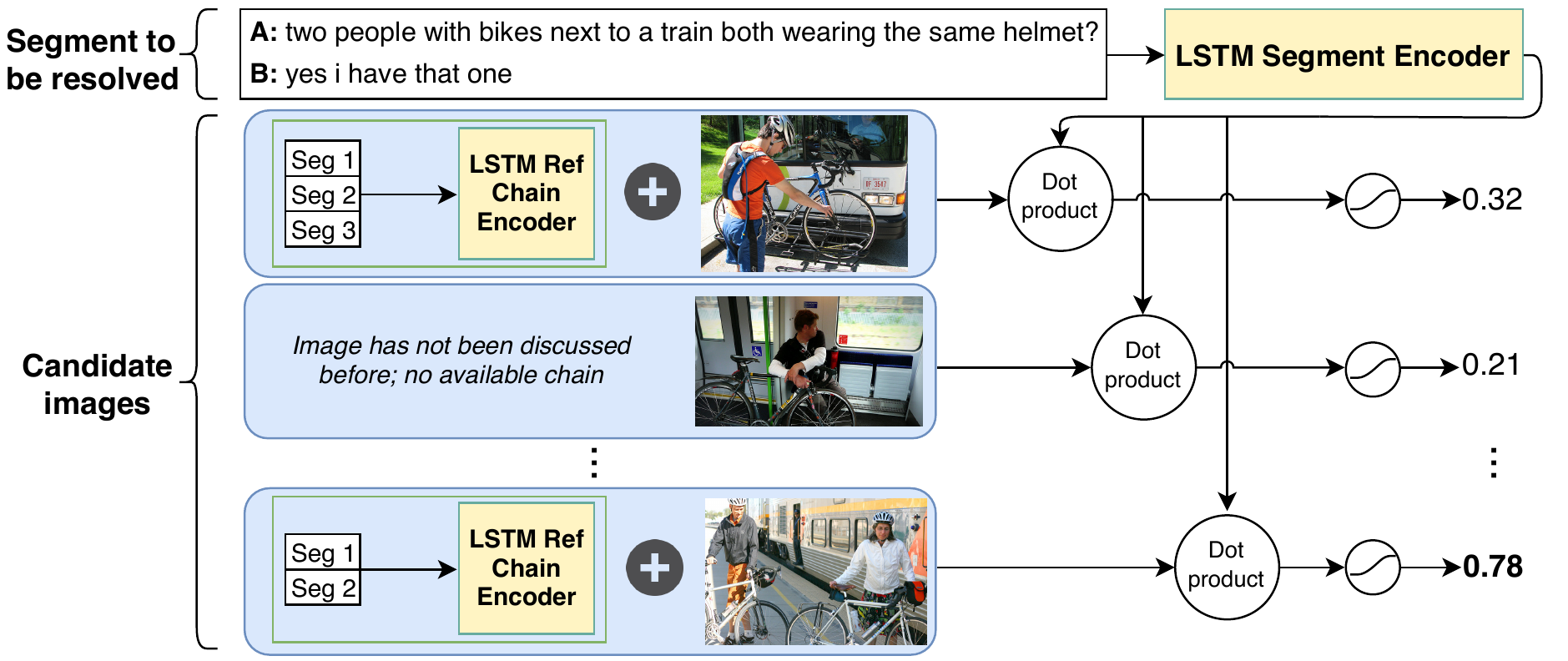}
\caption{Diagram of the model in the {\sc History} condition. For simplicity, we only show three candidate images. Some candidate images may not have a reference chain associated with them, while others may be linked to chains of different length, reflecting how many times an image has been referred to in the dialogue so far. In this example, the model predicts that the bottom candidate is the target referent of the segment to be resolved.}
\label{fig:model}
\end{figure*}

\subsection{Models}

Our resolution model encodes the linguistic features of the dialogue segment to be resolved 
with a recurrent neural network with Long Short-Term Memory \cite[LSTM,][]{Hochreiter1997}.
The last hidden state of the LSTM is then used as the representation for the dialogue segment. 
For each candidate image in the context, we obtain image features using the activations from the penultimate layer of a ResNet-152 \cite{He2015} pre-trained on ImageNet \cite{Deng2009}. These image features, which are of size 2048, are projected onto a smaller dimension equal to the hidden dimension of LSTM units. Projected image features go through ReLU non-linearity and are normalised to unit vectors. To assess which of the candidate images is a target, in the {\sc No-History} condition we take the dot product between the dialogue segment representation and each image feature vector, ending up with scalar predictions for all $N$ images in the context: $s = \{s_0, ... , s_N\}$.

For the {\sc History} condition, we propose 
a simple mechanism for taking into account linguistic common ground about each image. 
For each candidate image, 
we consider the sequence of previous segments within its reference chain. This shared linguistic background is encoded with another LSTM, whose last hidden state is added to the corresponding image feature that was projected to the same dimension as the hidden state of the LSTM. The resulting representation goes through ReLU, is normalised, and compared to the target dialogue segment representation via dot product, as in {\sc No-History} (see Figure~\ref{fig:model}).

As an ablation study, we train a {\sc History} model without visual features. This allows us to establish a baseline performance only involving language and to study whether the {\sc History} model with visual features learns an efficient multi-modal representation. We hypothesise that some descriptions can be successfully resolved by just comparing the current segment and the reference chain in the history (e.g., when descriptions are detailed and repeated). However, performance should be significantly lower than with visual features, for example when referring expressions are ambiguous. 

Sigmoid is applied element-wise over the scalar predictions in all three models. As a result, each image can be assessed independently using a decision threshold (set to $0.5$). This allows the model to predict multiple images as referents.\footnote{As explained in Section~\ref{sec:chains}, 25\% of segments are linked to two targets; 3\% to more. See Appendix~\ref{app:chains} for further details.}
We use Binary Cross Entropy Loss to train the models. Since distractor images make up $84.74\%$ of the items to be classified in the training set and target images constitute only the $15.26\%$ of them, we provided $84.74/15.26 \approx 5.5$ as the weight of the target class in the loss function.

All models were implemented in PyTorch, trained with a learning rate of 0.001 and a batch size of 512. The dimension of the word embeddings and the hidden dimensions of the LSTM units were all set to 512. The parameters were optimised using Adam \cite{DBLP:journals/corr/KingmaB14}. The models were trained until the validation loss stopped improving, after which we selected 
the model with the best weighted average of the target and non-target F-scores.


\subsection{Results}

We report precision, recall, and F-score for the target images in Table~\ref{tab:results}. 
Results for non-target images are available in Appendix~\ref{app:exp}. 
Every candidate image contributes individually to the scores, i.e., the task is not treated as multi-label for evaluation purposes.
Random baseline scores are obtained by taking the average of 10 runs with a model that predicts targets and non-targets randomly for the images in the test set. 

Given the low ratio of target to distractor images (see Table~\ref{tab:splits} in Section~\ref{sec:exp-data}),
the task of identifying target images 
is challenging and the random baseline achieves an F-score below 30\%. 
The results show that the resolution capabilities of our model are well above the baseline. 
The {\sc History} model achieves higher recall and F-score than the {\sc No-History} model, while precision is comparable across these two conditions.

\begin{table}[ht]\centering
\scalebox{0.9}{\begin{tabular}{|l|c|c|c|}
\hline
\textbf{Model}   & \multicolumn{1}{l|}{\textbf{Precision}} & \bf Recall & \bf F1 \\ \hline
Random baseline & 15.34 & 49.95 & 23.47\\ \hline
{\sc No-History} & 56.65 & 75.86 & 64.86\\ \hline
{\sc History}       & 56.66 & 77.41 & 65.43 \\ \hline
{\sc History} $\!$/$\!$ No image & 35.66	 & 63.18 & 45.59\\\hline
\end{tabular}}
\caption{Results for the target images in the test set.}
\label{tab:results}
\end{table}

\begin{figure}[ht]
\includegraphics[width=\columnwidth]{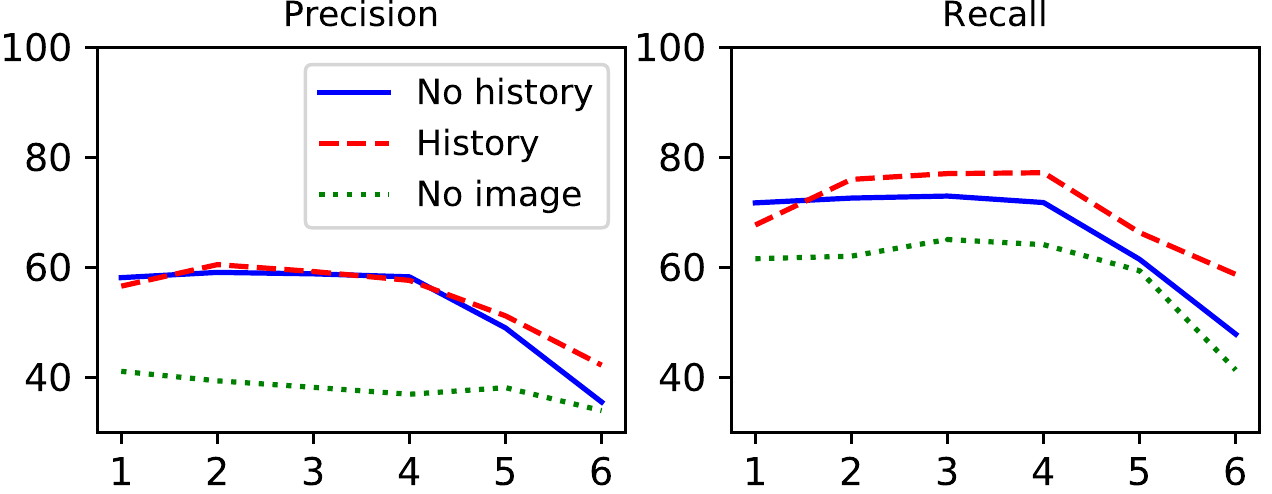}
\caption{Precision and recall ($y$ axis) for target images, given the position of the segment in a reference chain ($x$ axis).}
\label{fig:results}
\end{figure}

\noindent
For a more in-depth analysis of the results, we examine how precision and recall vary depending on the position of the to-be-resolved segment within a reference chain. Figure~\ref{fig:results} displays this information. 
As hypothesised, we observe that resolution performance is lower for later segments in a reference chain.  
For example, while precision is close to 60\% for first mentions (position 1 in a chain), it declines by around 20 points for last mentions.

\noindent
The plots in Figure~\ref{fig:results} also show the impact of taking into account the common ground accumulated over a reference chain. This is most prominent with regard to the recall of target images. The {\sc History} model yields higher results than the {\sc No-History} model when it comes to resolving segments that refer to an image that has already been referred to earlier within the dialogue (positions $> 1$).
Yet, the presence of linguistic context does not fully cancel out the effect observed above: The performance of the {\sc History} model also declines for later segments in a chain, 
indicating that more sophisticated methods are needed to fully exploit shared linguistic information.

Experiments with the {\sc History} model without visual features ({\sc History}/No image) confirm our hypothesis. The {\sc History} model outperforms the ``blind''  model by about 21 points in precision and 14 points in recall. We thus conclude that even our simple fusion mechanism already allows for learning an efficient multimodal encoding and resolution of referring expressions.

\subsection{Qualitative Analysis}
\label{sec:qualitative}

\begin{figure*}[ht]
\includegraphics[height=2.61cm]{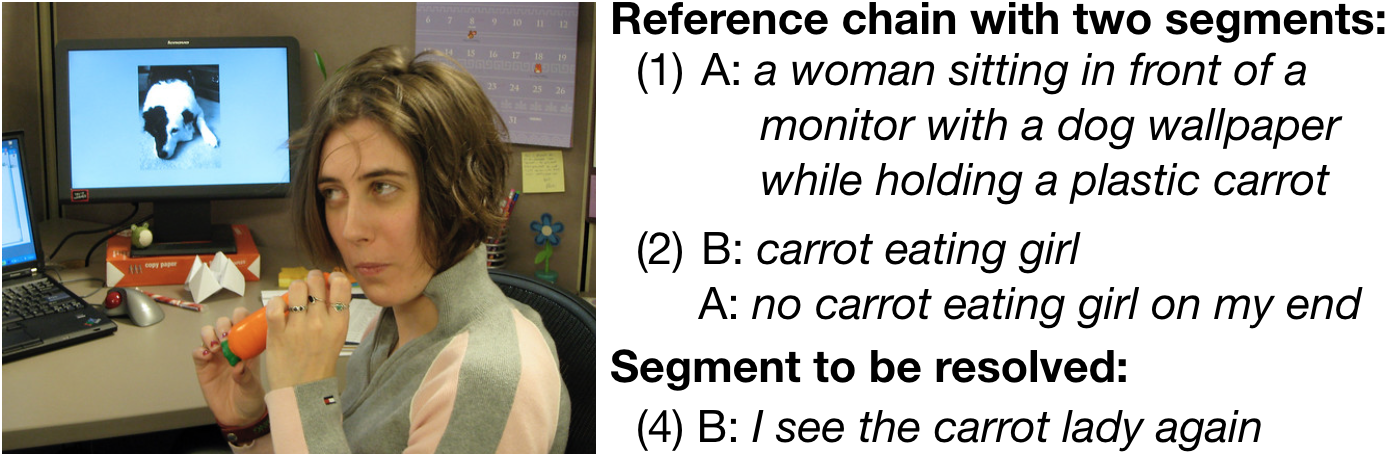}
\includegraphics[height=2.61cm]{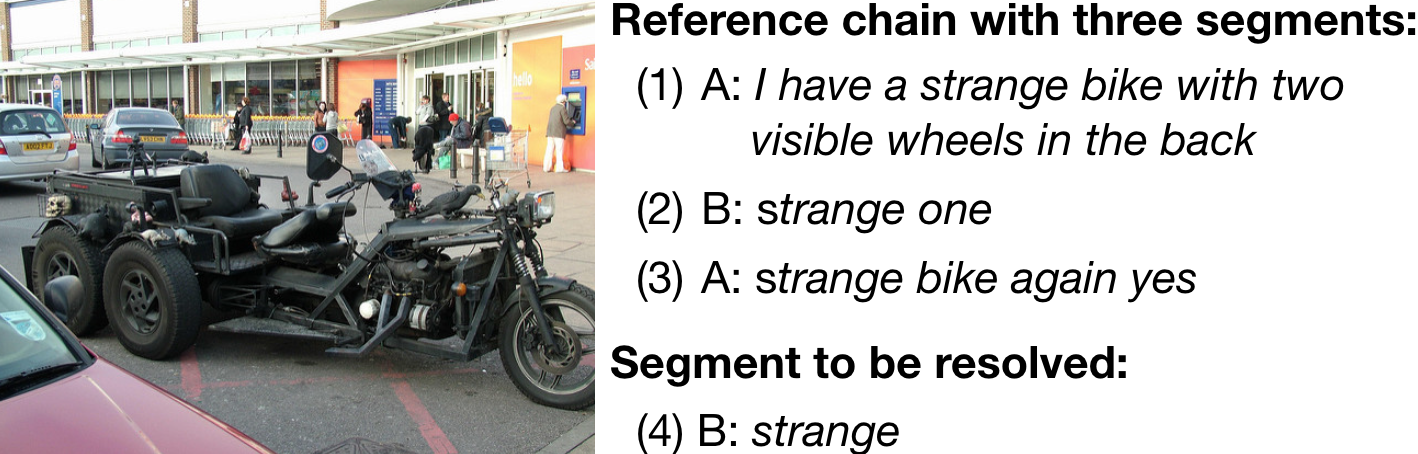} 
\caption{Reference chain for each of the two displayed images. The dialogue segments in the chains are slightly simplified for space reasons. \textbf{Left:} Both the {\sc History} and the {\sc No-History} models succeed at identifying this image as the target of the segment to be resolved. \textbf{Right:} The {\sc No-History} model fails to recognise this image as the target of the segment to be resolved, while the {\sc History} model succeeds. 
The distractor images for these two examples are available in Appendix~\ref{app:exp}.}
\label{fig:ex-history}
\end{figure*}

The quantitative dataset analysis presented in Section~\ref{sec:data} showed that 
referring expressions become shorter over time, with
interlocutors being most likely to retain nouns and adjectives. 
Qualitative inspection of the reference chains reveals that this compression process can
lead to very non-standard descriptions.  
We hypothesise that the degree to which the compressed descriptions rely on visual information
has an impact on the performance of the models. 
For example, the {\sc No-History} model can be effective when the participants 
converge on a non-standard description which highlights a visual property of the 
target image that clearly discriminates it from the distractors. This is the case in the
example shown on the left-hand side of Figure~\ref{fig:ex-history}. The target image
shows a woman holding what seems to be a plastic carrot. This feature stands out in a
domain where all the candidate images include a person and a TV.\footnote{The COCO annotations for this image seem to be slightly off, as the image is tagged as including a TV but in fact shows a computer monitor.}
After an initial, longer description (\emph{`a woman sitting in front of a monitor with a dog wallpaper while holding a plastic carrot'}), the participants use the much more compact description \textit{`the carrot lady'}. 
Arguably, given the saliency of the carrot in the given context, relying on the preceding linguistic history is not
critical in this case, and thus both the {\sc No-History} and the {\sc History} model succeed in identifying the target.

We observe that the {\sc History} model is particularly helpful when the participants converge on a non-standard
description of a target image that cannot easily be grounded on visual information. 
The image and reference chain on the right-hand side of Figure~\ref{fig:ex-history} illustrate this point, 
where the description to be resolved is the remarkably abstract \textit{`strange'}. Here the 
 {\sc History} model succeeds while the {\sc No-History} model fails.
As in the previous example, the referring expression in the first segment of the reference chain for this image ({\em `a strange bike with two visible wheels in the back'}) includes more
descriptive content -- indeed, it is similar to a caption, as shown by our analysis in Section~\ref{sec:data-chains}.
By exploiting shared linguistic context, the {\sc History} model can not only interpret the 
non-standard phrase, but also recover additional properties of the image
not explicit in the segment to be resolved, which presumably help to ground it.

\section{Conclusion}

We have presented the first large-scale dataset of goal-oriented, visually grounded dialogues
for investigating shared linguistic history.
Through the data collection's task setup, participants repeatedly refer to a controlled set of target images, which allows them to improve task efficiency if they utilise their developing common ground and establish conceptual pacts \cite{Brennan1996} on referring expressions. 
The collected dialogues exhibit a significant shortening of utterances throughout a game, with final referring expressions starkly differing from both standard image captions and initial descriptions. 
To illustrate the potential of the dataset, we trained a baseline reference resolution model and showed that information accumulated over a reference chain helps to resolve later descriptions. Our results suggest that more sophisticated models are needed to fully exploit shared linguistic history. 

The current paper showcases only some of the aspects of the PhotoBook dataset, which we hereby release to the public (\url{https://dmg-photobook.github.io}). In future work, the data can be used to further investigate common ground and conceptual pacts; be extended through manual annotations for a more thorough linguistic analysis of co-reference chains; exploit the combination of vision and language to develop computational models for referring expression generation; or use the PhotoBook task in the ParlAI framework for Turing-Test-like evaluation of  dialogue agents. 


\section*{Acknowledgements}

The PhotoBook dataset was collected thanks to funding 
in the form of a Facebook ParlAI Research Award to Raquel Fern\'andez. We are grateful to the ParlAI team, in particular
Jack Urbanek and Jason Weston, for their continued support and assistance in using the ParlAI framework during the data collection, which was coordinated by Janosch Haber. 
We warmly thank the volunteers who took part in pilot experiments. 
Thanks also go to Aashish Venkatesh for his valuable contributions in brainstorming sessions about the task design and for preliminary modelling efforts. 
Finally, we are grateful to the participants of two Faculty Summits at Facebook AI Research NYC for their feedback.

\bibliography{bibliography}

\begin{thebibliography}{32}
\expandafter\ifx\csname natexlab\endcsname\relax\def\natexlab#1{#1}\fi

\bibitem[{von Ahn et~al.(2006)von Ahn, Kedia, and Blum}]{vonAhn2006}
Luis von Ahn, Mihir Kedia, and Manuel Blum. 2006.
\newblock \href {https://doi.org/10.1145/1124772.1124784} {Verbosity: A game
  for collecting common-sense facts}.
\newblock In \emph{Proceedings of the Conference on Human Factors in Computing
  Systems}.

\bibitem[{Anderson et~al.(1991)Anderson, Bader, Bard, Boyle, Doherty, Garrod,
  Isard, Kowtko, McAllister, Miller et~al.}]{anderson1991hcrc}
Anne~H. Anderson, Miles Bader, Ellen~Gurman Bard, Elizabeth Boyle, Gwyneth
  Doherty, Simon Garrod, Stephen Isard, Jacqueline Kowtko, Jan McAllister, Jim
  Miller, et~al. 1991.
\newblock {The HCRC Map Task corpus}.
\newblock \emph{Language and Speech}, 34(4):351--366.

\bibitem[{Antol et~al.(2015)Antol, Agrawal, Lu, Mitchell, Batra, Zitnick, and
  Parikh}]{Antol2015}
Stanislaw Antol, Aishwarya Agrawal, Jiasen Lu, Margaret Mitchell, Dhruv Batra,
  C.~Lawrence Zitnick, and Devi Parikh. 2015.
\newblock {VQA: Visual Question Answering}.
\newblock In \emph{Proceedings of ICCV}.

\bibitem[{Brennan and Clark(1996)}]{Brennan1996}
Susan~E. Brennan and Herbert~H. Clark. 1996.
\newblock Conceptual pacts and lexical choice in conversation.
\newblock \emph{Journal of Experimental Psychology: Learning, Memory, and
  Cognition}, 22:1482--1493.

\bibitem[{Brown-Schmidt et~al.(2015)Brown-Schmidt, Yoon, and
  Ryskin}]{brown2015people}
Sarah Brown-Schmidt, Si~On Yoon, and Rachel~Anna Ryskin. 2015.
\newblock People as contexts in conversation.
\newblock In \emph{Psychology of Learning and Motivation}, volume~62,
  chapter~3, pages 59--99. Elsevier.

\bibitem[{Clark(1996)}]{Clark1996}
Herbert~H. Clark. 1996.
\newblock \href {https://doi.org/10.1017/CBO9780511620539} {\emph{Using
  Language}}.
\newblock 'Using' Linguistic Books. Cambridge University Press.

\bibitem[{Clark and Wilkes-Gibbs(1986)}]{Clark1986}
Herbert~H. Clark and Deanna Wilkes-Gibbs. 1986.
\newblock \href {https://doi.org/https://doi.org/10.1016/0010-0277(86)90010-7}
  {Referring as a collaborative process}.
\newblock \emph{Cognition}, 22(1):1 -- 39.

\bibitem[{Das et~al.(2017)Das, Kottur, Gupta, Singh, Yadav, Moura, Parikh, and
  Batra}]{Das2016}
Abhishek Das, Satwik Kottur, Khushi Gupta, Avi Singh, Deshraj Yadav,
  Jos\'e~M.F. Moura, Devi Parikh, and Dhruv Batra. 2017.
\newblock {V}isual {D}ialog.
\newblock In \emph{Proceedings of CVPR}.

\bibitem[{De~Vries et~al.(2017{\natexlab{a}})De~Vries, Strub, Chandar,
  Pietquin, Larochelle, and Courville}]{deVries2016}
Harm De~Vries, Florian Strub, Sarath Chandar, Olivier Pietquin, Hugo
  Larochelle, and Aaron Courville. 2017{\natexlab{a}}.
\newblock Guesswhat?! visual object discovery through multi-modal dialogue.
\newblock In \emph{Proc. of CVPR}.

\bibitem[{De~Vries et~al.(2017{\natexlab{b}})De~Vries, Strub, Mary, Larochelle,
  Pietquin, and Courville}]{deVries2017}
Harm De~Vries, Florian Strub, J{\'e}r{\'e}mie Mary, Hugo Larochelle, Olivier
  Pietquin, and Aaron Courville. 2017{\natexlab{b}}.
\newblock \href {https://hal.inria.fr/hal-01648683} {{Modulating early visual
  processing by language}}.
\newblock In \emph{{Proceedings of NIPS}}.

\bibitem[{Deng et~al.(2009)Deng, Dong, Socher, Li, Li, and Fei-Fei}]{Deng2009}
Jia Deng, Wei Dong, Richard Socher, Li-Jia Li, Kai Li, and Li~Fei-Fei. 2009.
\newblock {ImageNet: A Large-Scale Hierarchical Image Database}.
\newblock In \emph{Proc.~of CVPR}.

\bibitem[{Hara et~al.(2018)Hara, Adams, Milland, Savage, Callison-Burch, and
  Bigham}]{Hara2017}
Kotaro Hara, Abigail Adams, Kristy Milland, Saiph Savage, Chris Callison-Burch,
  and Jeffrey~P Bigham. 2018.
\newblock {A Data-Driven Analysis of Workers' Earnings on Amazon Mechanical
  Turk}.
\newblock In \emph{Proceedings of the Conference on Human Factors in Computing
  Systems}.

\bibitem[{He et~al.(2017)He, Balakrishnan, Eric, and Liang}]{He2017}
He~He, Anusha Balakrishnan, Mihail Eric, and Percy Liang. 2017.
\newblock \href {http://arxiv.org/abs/1704.07130} {Learning symmetric
  collaborative dialogue agents with dynamic knowledge graph embeddings}.
\newblock In \emph{Proceedings of ACL}.

\bibitem[{He et~al.(2016)He, Zhang, Ren, and Sun}]{He2015}
Kaiming He, Xiangyu Zhang, Shaoqing Ren, and Jian Sun. 2016.
\newblock Deep residual learning for image recognition.
\newblock In \emph{Proceedings of CVPR}.

\bibitem[{Hochreiter and Schmidhuber(1997)}]{Hochreiter1997}
Sepp Hochreiter and J\"{u}rgen Schmidhuber. 1997.
\newblock \href {https://doi.org/10.1162/neco.1997.9.8.1735} {Long short-term
  memory}.
\newblock \emph{Neural Computation}, 9(8):1735--1780.

\bibitem[{Ilinykh et~al.(2018)Ilinykh, Zarrie{\ss}, and
  Schlangen}]{ilinykh2018task}
Nikolai Ilinykh, Sina Zarrie{\ss}, and David Schlangen. 2018.
\newblock The task matters: Comparing image captioning and task-based
  dialogical image description.
\newblock In \emph{Proceedings of INLG}.

\bibitem[{Kazemzadeh et~al.(2014)Kazemzadeh, Ordonez, Matten, and
  Berg}]{Kazemzadeh2014ReferIt}
Sahar Kazemzadeh, Vicente Ordonez, Mark Matten, and Tamara~L. Berg. 2014.
\newblock {ReferIt Game: Referring to Objects in Photographs of Natural
  Scenes}.
\newblock In \emph{Proceedings of EMNLP}.

\bibitem[{Kingma and Ba(2014)}]{DBLP:journals/corr/KingmaB14}
Diederik~P. Kingma and Jimmy Ba. 2014.
\newblock \href {http://arxiv.org/abs/1412.6980} {Adam: {A} method for
  stochastic optimization}.
\newblock \emph{CoRR}, abs/1412.6980.

\bibitem[{Krauss and Weinheimer(1964)}]{Krauss1964}
Robert~M. Krauss and Sidney Weinheimer. 1964.
\newblock \href {https://doi.org/10.3758/BF03342817} {Changes in reference
  phrases as a function of frequency of usage in social interaction: a
  preliminary study}.
\newblock \emph{Psychonomic Science}, 1(1):113--114.

\bibitem[{Krauss and Weinheimer(1966)}]{Krauss1966}
Robert~M. Krauss and Sidney Weinheimer. 1966.
\newblock \href {https://doi.org/10.1037/h0023705} {Concurrent feedback,
  confirmation, and the encoding of referents in verbal communication.}
\newblock \emph{Journal of Personality and Social Psychology}, 4(3):343--346.

\bibitem[{Krishna et~al.(2017)Krishna, Zhu, Groth, Johnson, Hata, Kravitz,
  Chen, Kalantidis, Li, Shamma et~al.}]{VisualGenome}
Ranjay Krishna, Yuke Zhu, Oliver Groth, Justin Johnson, Kenji Hata, Joshua
  Kravitz, Stephanie Chen, Yannis Kalantidis, Li-Jia Li, David~A Shamma, et~al.
  2017.
\newblock {Visual Genome: Connecting language and vision using crowdsourced
  dense image annotations}.
\newblock \emph{International Journal of Computer Vision}, 123(1):32--73.

\bibitem[{Likert(1932)}]{Likert1932}
Rensis Likert. 1932.
\newblock A technique for the measurement of attitudes.
\newblock \emph{Archives of Psychology}, 22 140:55--55.

\bibitem[{Lin et~al.(2014)Lin, Maire, Belongie, Hays, Perona, Ramanan,
  Doll{\'a}r, and Zitnick}]{LinCOCO2014}
Tsung-Yi Lin, Michael Maire, Serge Belongie, James Hays, Pietro Perona, Deva
  Ramanan, Piotr Doll{\'a}r, and C.~Lawrence Zitnick. 2014.
\newblock {Microsoft COCO: Common Objects in Context}.
\newblock In \emph{Proc.~of ECCV}.

\bibitem[{Mao et~al.(2016)Mao, Huang, Toshev, Camburu, Yuille, and
  Murphy}]{Mao2015}
Junhua Mao, Jonathan Huang, Alexander Toshev, Oana Camburu, Alan Yuille, and
  Kevin Murphy. 2016.
\newblock \href {https://doi.org/10.1109/CVPR.2016.9} {Generation and
  comprehension of unambiguous object descriptions}.
\newblock In \emph{Proceedings of CVPR}.

\bibitem[{Mikolov et~al.(2013)Mikolov, Sutskever, Chen, Corrado, and
  Dean}]{Mikolov2013}
Tomas Mikolov, Ilya Sutskever, Kai Chen, Greg Corrado, and Jeffrey Dean. 2013.
\newblock \href {http://dl.acm.org/citation.cfm?id=2999792.2999959}
  {Distributed representations of words and phrases and their
  compositionality}.
\newblock In \emph{{Proceedings of NIPS}}.

\bibitem[{Miller et~al.(2017)Miller, Feng, Batra, Bordes, Fisch, Lu, Parikh,
  and Weston}]{Miller2017}
Alexander Miller, Will Feng, Dhruv Batra, Antoine Bordes, Adam Fisch, Jiasen
  Lu, Devi Parikh, and Jason Weston. 2017.
\newblock \href {http://aclweb.org/anthology/D17-2014} {{ParlAI: A Dialog
  Research Software Platform}}.
\newblock In \emph{Proceedings of EMNLP}.

\bibitem[{Plummer et~al.(2015)Plummer, Wang, Cervantes, Caicedo, Hockenmaier,
  and Lazebnik}]{plummer2015flickr30k}
Bryan~A Plummer, Liwei Wang, Chris~M Cervantes, Juan~C Caicedo, Julia
  Hockenmaier, and Svetlana Lazebnik. 2015.
\newblock Flickr30k entities: Collecting region-to-phrase correspondences for
  richer image-to-sentence models.
\newblock In \emph{Proceedings of ICCV}.

\bibitem[{Stalnaker(1978)}]{Stalnaker1978}
Robert Stalnaker. 1978.
\newblock Assertion.
\newblock In P.~Cole, editor, \emph{Pragmatics}, volume~9 of \emph{Syntax and
  Semantics}, pages 315--332. New York Academic Press.

\bibitem[{Takenobu et~al.(2012)Takenobu, Ryu, Asuka, and
  Naoko}]{takenobu2012rex}
Tokunaga Takenobu, Iida Ryu, Terai Asuka, and Kuriyama Naoko. 2012.
\newblock {The REX corpora: A collection of multimodal corpora of referring
  expressions in collaborative problem solving dialogues}.
\newblock In \emph{Proceedings of LREC}.

\bibitem[{Vedantam et~al.(2017)Vedantam, Bengio, Murphy, Parikh, and
  Chechik}]{Vedantam2017}
Ramakrishna Vedantam, Samy Bengio, Kevin Murphy, Devi Parikh, and Gal Chechik.
  2017.
\newblock \href {https://doi.org/10.1109/CVPR.2017.120} {Context-aware captions
  from context-agnostic supervision}.
\newblock In \emph{Proceedings of CVPR}.

\bibitem[{Yoshida(2011)}]{yoshida2011referring}
Etsuko Yoshida. 2011.
\newblock \emph{Referring expressions in English and Japanese: patterns of use
  in dialogue processing}, volume 208.
\newblock John Benjamins Publishing.

\bibitem[{Zarrie{\ss} et~al.(2016)Zarrie{\ss}, Hough, Kennington,
  Manuvinakurike, DeVault, Fern{\'a}ndez, and Schlangen}]{Zarriess2016}
Sina Zarrie{\ss}, Julian Hough, Casey Kennington, Ramesh Manuvinakurike, David
  DeVault, Raquel Fern{\'a}ndez, and David Schlangen. 2016.
\newblock \href
  {http://www.lrec-conf.org/proceedings/lrec2016/pdf/563_Paper.pdf}
  {{PentoRef}: {A} {Corpus} of {Spoken} {References} in {Task}-oriented
  {Dialogues}}.
\newblock In \emph{Proceedings of LREC}.

\end{thebibliography}
\bibliographystyle{acl_natbib}

\section*{Appendices}
\appendix

\section{Task Setup}
\label{app:task}

\paragraph{Image Sets}

The images used in the PhotoBook task are taken from the MS COCO 2014 Trainset \cite{LinCOCO2014}. Images in MS COCO were collected from the Flickr\footnote{\url{https://www.flickr.com/}} image repository, which contains labelled photos predominantly uploaded by amateur photographers. The pictures largely are snapshots of everyday situations, placing objects in a natural and often rich context (hence the name \textit{Common Objects in COntext}) instead of showing an iconic view of objects. In the MS COCO Trainset, images are manually annotated with the outlines of the depicted objects as well as their object categories. We use this information to select similar pictures to display in the PhotoBook task.
Through the filtering described in Section~\ref{sec:task}, we obtained 30 sets of similar images with different pairings of their most prominent objects. In total there are 26 unique object categories in the image sets. 
The most frequent object category in the image sets is \texttt{person}, which is one of the two main objects in 19 sets.

\paragraph{Specification of Games}

We developed a simple function to select which images of a set should be shown to which participant in which round of a game in order to guarantee that the task setup elicits sufficient image (re-)references for collecting co-reference chains. In this function, the 12 images in a set of similar photographs are randomly indexed and then assigned to a participant's display based on the schema displayed in Table \ref{tab:image_sets}. With this schema, each photograph is displayed exactly five times while the order of images and the order of rounds can be randomised to prevent participants from detecting patterns in the display. 
Each of these sets then is duplicated and assigned a different selection of highlighted images to obtain the 60 game sets of the PhotoBook task.  While most highlighted images recur five times during a game, they can be highlighted for both participants in the same round. As a result, any given image is highlighted in an average of 3.42 rounds of a game (see Table \ref{tab:highlighting_schema} for the highlighting schema).

\begin{table}[h]
    \centering
    \scalebox{0.8}{
    \begin{tabular}{|c|l|l|}
    \hline 
    \textbf{Round} & \textbf{Participant A} & \textbf{Participant B} \\
    \hline
        1 & 1, 2, 3, 4, 5, 6  & 1, 2, 3, 4, 7, 8  \\
        2 & 1, 3, 6, 7, 9, 10  & 2, 3, 6, 7, 9, 11 \\
        3 & 4, 5, 7, 10, 11, 12 & 2, 3, 6, 7, 9, 11 \\
        4 & 1, 2, 5, 8, 11, 12 & 1, 4, 5, 8, 10, 12 \\
        5 & 5, 6, 8, 9, 11, 12 & 3, 7, 9, 10, 11, 12\\
   \hline 
    \end{tabular}}
    \caption{Assignment of image IDs to the different participants and rounds of a game schema. The order of rounds and the arrangement of images on the participant's display can be randomised without effect on the game setup.}
    \label{tab:image_sets}
\end{table}

\begin{table}[h]
\centering
\scalebox{0.75}{\begin{tabular}{|r|cc|cc|cc|cc|cc|c|cc|c|}
\hline
   \multicolumn{11}{|c}{\hspace{3em}\textbf{Game Round}} & \multicolumn{4}{|c|}{\textbf{Statistics}} \\
   \hline
   & \multicolumn{2}{c}{\textbf{1}} & \multicolumn{2}{c}{\textbf{2}} & \multicolumn{2}{c}{\textbf{3}} & \multicolumn{2}{c}{\textbf{4}} & \multicolumn{2}{c|}{\textbf{5}} & & \multicolumn{2}{c|}{\textbf{H}} &  \\
   \textbf{ID}  & \textbf{A} & \textbf{B} & \textbf{A} & \textbf{B} & \textbf{A} & \textbf{B} & \textbf{A} & \textbf{B} & \textbf{A} & \textbf{B} &  \textbf{T} & \textbf{1} & \textbf{2} & \textbf{R} \\
   \hline
1  & 1         & 1         & 1         &           &           &           & 1         & 1         &           &           & 5     & 5              & 0              & 3      \\
2  & 2         & 2         &           & 2         &           & 2         & 2         &           &           &           & 5     & 0              & 5              & 4      \\
3  & 1         & 1         & 1         & 1         &           &           &           &           &           & 1         & 5     & 5              & 0              & 3      \\
4  & 2         & 2         &           &           & 2         & 1         &           & 2         &           &           & 5     & 4              & 1              & 3      \\
5  & 1         &           &           &           & 1         &           & 1         & 1         & 1         &           & 5     & 0              & 5              & 4      \\
6  & 2         &           & 2         & 2         &           & 2         &           &           & 2         &           & 5     & 5              & 0              & 4      \\
7  &           & 1         & 1         & 1         & 1         &           &           &           &           & 1         & 5     & 0              & 5              & 3      \\
8  &           & 2         &           &           &           & 1         & 2         & 1         & 1         &           & 5     & 2              & 3              & 3      \\
9  &           &           & 2         & 2         &           & 1         &           &           & 2         & 2         & 5     & 4              & 1              & 3      \\
10 &           &           & 2         &           & 2         & 2         &           & 2         &           & 2         & 5     & 5              & 0              & 4      \\
11 &           &           &           & 1         & 1         &           & 1         &           & 1         & 1         & 5     & 0              & 5              & 4      \\
12 &           &           &           &           & 2         &           & 2         & 2         & 2         & 2         & 5     & 5              & 0              & 3     \\
\hline
\end{tabular}}
\caption{Schema of referent image highlighting in the PhotoBook task. The left part of the table indicates whether a given image is highlighted for one of the two participants (\textbf{A} and \textbf{B}) in a given game round in either game 1 or 2. \textbf{T} indicates the total count of highlights (which is 5 always), \textbf{H} counts the highlights per game and \textbf{R} the number of rounds that an image is highlighted in.}
\label{tab:highlighting_schema}
\end{table}

\section{Task Instructions}
\label{app:collection}

\paragraph{HIT Preview} \label{app:hit_preview}

\begin{figure*}[h]
\includegraphics[width=\linewidth]{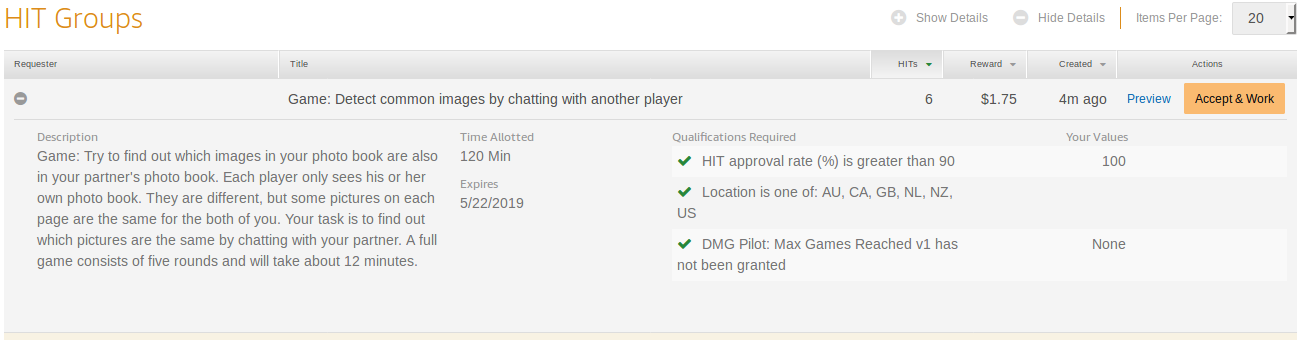}
\caption{Screenshot of the PhotoBook task AMT HIT details shown to a participant.}
\label{fig:hit_description}
\end{figure*}

\begin{figure*}[h]
\includegraphics[width=\linewidth]{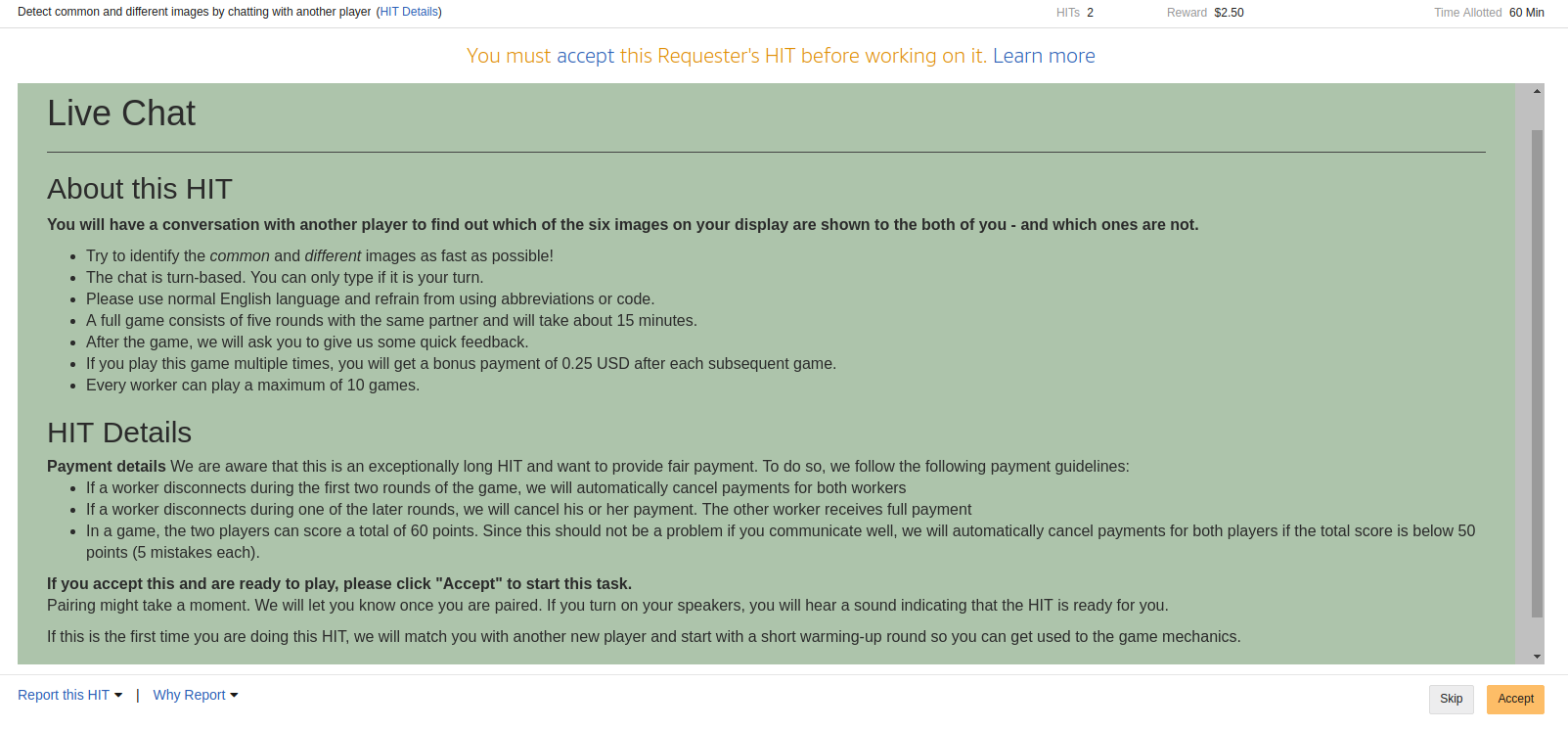}

\vspace*{-10pt}
\caption{Screenshot of the PhotoBook task AMT HIT preview page.}
\label{fig:preview_page}
\end{figure*}

When the PhotoBook task environment is initialised, it publishes a specified number of Human Intelligence Tasks (HITs) titled \textit{Game: Detect common images by chatting with another player} on Amazon Mechanical Turk (see Figure \ref{fig:hit_description} for a full print of the descriptions). Participants entering the HIT are shown a preview screen with the central task details as shown in Figure \ref{fig:preview_page}.

\paragraph{Game Round Mechanics} 

\begin{figure*}[ht]
     \subfloat[Screenshot of the PhotoBook task's display for one of the participants during the warming-up round.\label{fig:warmup_screenshot}]{%
       \includegraphics[width=0.48\textwidth]{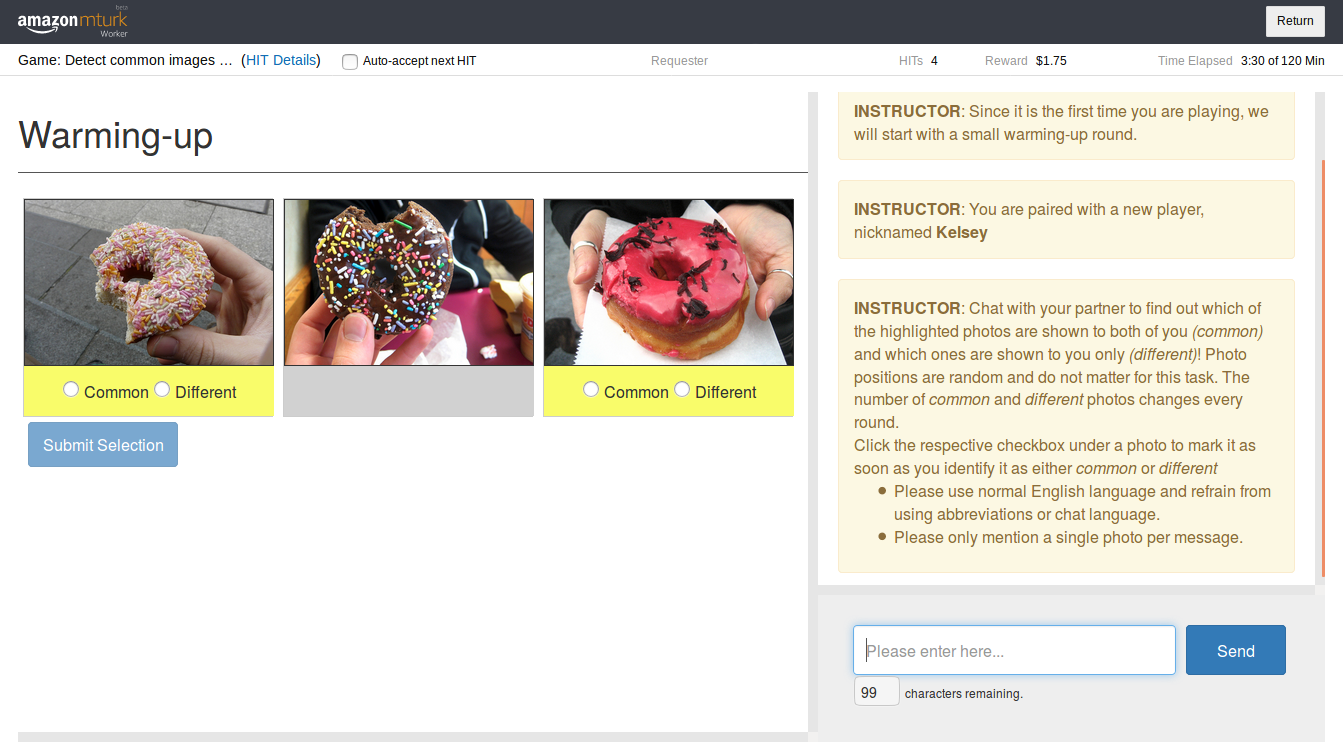}
     }
     \hfill
     \subfloat[Example screenshot of the PhotoBook AMT feedback display. \label{fig:feedback_interface}]{%
       \includegraphics[width=0.48\textwidth]{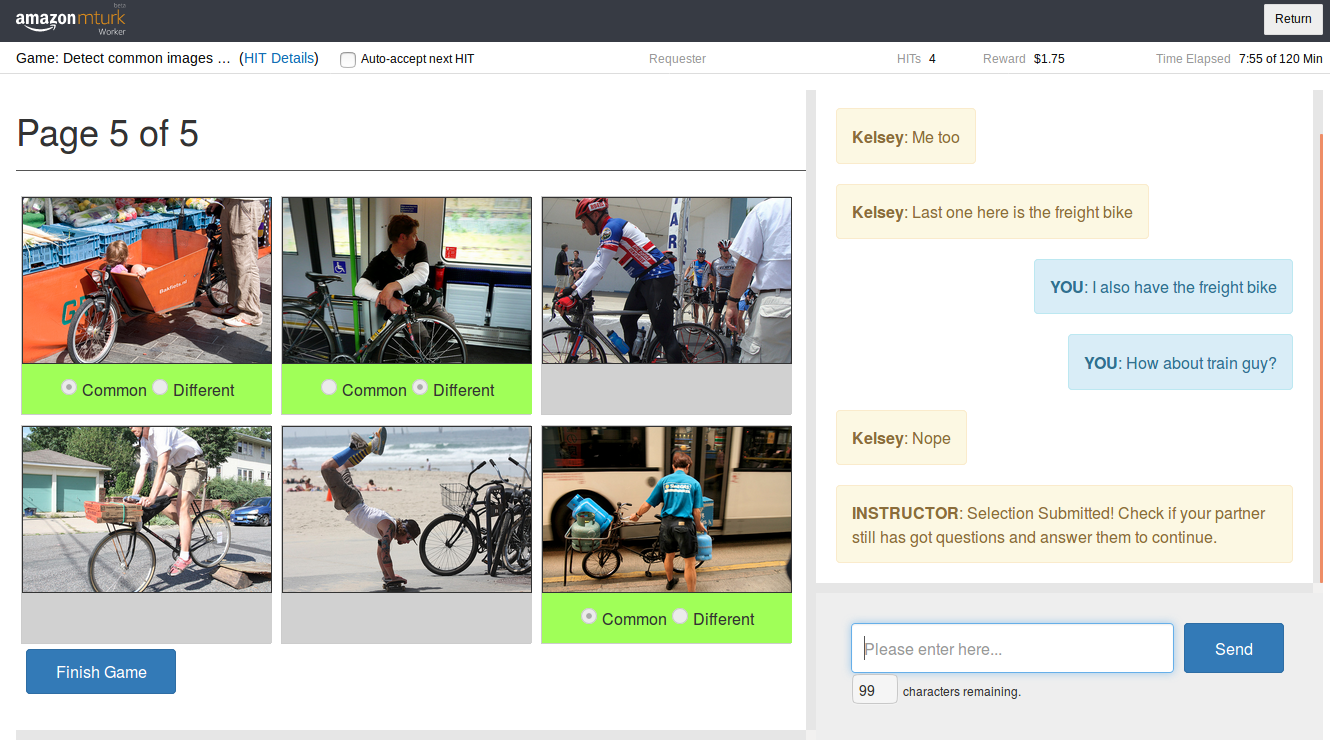}
     }
     \caption{Example screenshots for a participant's display during the warming-up round and feedback screen.}
     \label{fig:game_interfaces}
   \end{figure*}

The PhotoBook task AMT user interface is designed in such a way that the six images per round are displayed in a 2 $\times$ 3 grid, with a coloured bar under each image: If the image is highlighted, this bar is yellow and contains a radio button option for the \textit{common} and \textit{different} labels. If they are not highlighted for a player, the bar is greyed out and empty. The submit button is deactivated as long as not all highlighted images have been labelled. As soon as both players submitted their selection, a feedback page is shown where the bars under the highlighted images either colour green to indicate a correct selection or red to indicate a wrong one. Figure \ref{fig:game_interfaces}\subref{fig:feedback_interface} shows a screenshot of the feedback display. 

The radio buttons are disabled in the feedback screens so players cannot revise their selection - they can however communicate about their mistakes or pass any other feedback to their partner. The title of a page indicates the current page number so participants can always check their progress; the text input field is limited to a maximum of hundred characters to prevent listings of multiple images or overly elaborate descriptions - which, if necessary, can be conveyed in a number of subsequent messages.

\paragraph{Feedback Questionnaire} 

In order to facilitate a qualitative analysis of dialogue agents developed for the PhotoBook task, we also collect a gold-standard benchmark of participant's self-reported satisfaction scores. These scores later can be compared with those obtained by pairing human participants with an artificial dialogue agent in order to assess it in a Turing Test-like setting. Following \citet{He2017}, we ask participants to rate three statements on a five-point Likert scale \cite{Likert1932}, ranging from \textit{strongly agree} to \textit{strongly disagree}:
\begin{enumerate}[itemsep=0pt,leftmargin=15pt]
    \item Overall collaboration with my partner worked well.
    \item I understood my partner's descriptions well. 
    \item My partner seemed to understand me well.
\end{enumerate}

\paragraph{Warming-Up Round} 

During an initial series of pilot runs we observed that for new participants the first round of their first game took significantly longer than any other ones. Although we do expect that participants get more efficient over time, we argue that this effect is largely related to the fact that participants need to get familiar with the task's mechanics when it is the first time they are exposed to it. In order to control for this effect, we added a warming-up round with three images per participant\footnote{Warming-Up image categories are disjoint from the regular PhotoBook image sets.} for each pair of new participants (see Figure \ref{fig:warmup_screenshot}). This strongly reduced the completion time of new participants' first game rounds.

\paragraph{Matching Participants} \label{app:pairing}
In order to collect unbiased samples of the referring expression generation process, we aim to prevent both, i) participants completing the same game multiple times (as here they could re-use referring expressions that worked well during the last one) and ii) specific pairs of participants completing multiple games (as they might have established some kind of strategy or code already). We however also aim at designing the task in such a way that the degree of the partner-specificity in established canonical expressions could be assessed. To achieve this, the participant matching should create settings where a re-entering participant is assigned a game with the same image set as in the game before, but paired with a different conversation partner changes \cite[compare for example][]{Brennan1996}. In order to maximise the number of this second game setting, we encourage workers to continue playing by paying them a bonus of 0.25 USD for each 2nd, 3rd, 4th and 5th game. 

\paragraph{Worker Payment} 
The HIT description also details the worker's payment. We want to provide fair payment to workers, which we calculated based on an average wage of 10 USD per hour \cite{Hara2017}.\footnote{See also \href{http://wiki.wearedynamo.org/index.php?title=Fair_payment}{DynamoWiki}.} 
An initial set of runs resulted in an average completion time of 12 minutes, which indicated an expected expense of about 2 USD per participant per game. More experienced workers however managed to complete a full game in six to ten minutes, meaning that for them we would often surpass the 10 USD/h guideline based on this calculation. Other workers -- especially new ones - took up to 25 minutes for the first game, which means that they on the other hand would be strongly under-payed with a rigid per-game payment strategy. To mitigate this effect, we developed the following payment schema: Each worker that completes a full game is payed a base-amount of 1.75 USD -- which is indicated in the HIT description. If the game took longer than ten minutes, the participants are payed a bonus amount of 0.10 USD per minute, up to an additional bonus payment of 1.50 USD for 25 or more minutes. In order to not encourage workers to play slowly, we only inform them about this bonus at the end of a HIT. With this bonus and the 20\% AMT fee on each transaction, we expected an average cost of about 5 USD per game, which due to connection problems in the framework ultimately accumulated to 6 USD for a completed game.
The total cost of the data collection, including pilot runs, was 16,350 USD.

\section{Dataset Samples}
\label{app:data}

Through the goal-oriented nature of participants' interactions in the PhotoBook dataset, we do not only collect image descriptions but rather the full, collaborative process of establishing, grounding and refining referring expressions throughout the  subsequent rounds of the PhotoBook task. As a result, we capture a wide range of dialogue acts such as clarification questions, corrections, extensions, (self-)repairs as well as interactions concerning game mechanics. Consider for example the following interactions: 

\begin{dialogue}
\speak{\textbf{A}} \textit{Man with dog on lap looking at his computer?} 
\speak{\textbf{B}} \textit{I don't have that, but could it be a TV in yours? Mine has a man sitting with his dog watching TV.} 
\speak{\textbf{A}} \textit{yes, TV - sorry!} 
\speak{\textbf{B}}\textit{Okay.}
\end{dialogue}

\vspace*{10pt}

\begin{dialogue}
\speak{\textbf{A}} \textit{Do you have someone on a big motorcycle and their head isn't visible?} 
\speak{\textbf{A}}\textit{There is a blue car in the background}
\speak{\textbf{B}} \textit{No}
\speak{\textbf{A}}\textit{In any of the pictures?}
\speak{\textbf{B}} \textit{No}
\speak{\textbf{A}} \textit{Okay, thank you}
\end{dialogue}

\pagebreak 

\begin{dialogue}
\speak{\textbf{B}} \textit{Woman with hot dog}
\speak{\textbf{A}} \textit{Older girl with glasses holding a hot dog?}
\speak{\textbf{B}} \textit{sitting}
\speak{\textbf{A}} \textit{Yeah}
\end{dialogue}

\begin{dialogue}
\speak{\textbf{A}} \textit{Do you have a picture with a lady in a fancy dress standing by a motorcycle?}
\speak{\textbf{B}} \textit{no}
\speak{\textbf{B}}\textit{wait}
\speak{\textbf{B}} \textit{yes, in black?}
\speak{\textbf{A}} \textit{Yes, it's a black dress with white trim.}
\end{dialogue}

\begin{dialogue}
\speak{\textbf{A}} \textit{Is there anything else?}
\speak{\textbf{B}}\textit{ Do you have the old lady in the white hat/blue pants reading?}
\speak{\textbf{A}} \textit{Yes, I do.}
\speak{\textbf{B}} \textit{Okay, that's all for me}
\end{dialogue}

\noindent In most cases, referring expressions agreed upon during the first rounds of a game are further refined and optimised while re-referring to the same target object in later rounds of the game. These refinements often are manifested in an omission of detail while retaining core features of the target object.

\begin{dialogue}
\speak{\textbf{A}}\textit{Do you have a boy with a teal coloured shirt with yellow holding a bear with a red shirt?} 
\speak{\textbf{B}} \textit{Yes} \\
--
\speak{\textbf{B}} \textit{Boy with teal shirt and bear with red shirt?} 
\speak{\textbf{A}} \textit{Yes!} \\
--
\speak{\textbf{A}} \textit{Teal shirt boy?} 
\speak{\textbf{B}} \textit{No} 
\end{dialogue}

\noindent Collecting all utterances that refer to a specific target image during a given game creates its co-reference chain. Consider the following examples of first (F) and last (L) referring expressions from co-reference chains manually extracted from the PhotoBook dataset:

\begin{dialogue}
\speak{\textbf{F}} \textit{Two girls near TV playing wii \\
One in white shirt, one in grey}
\speak{\textbf{L}} \textit{Girls in white and grey}
\end{dialogue}

\begin{dialogue}
\speak{\textbf{F}} \textit{A person that looks like a monk sitting on a bench \\
He's wearing a blue and white ball cap}
\speak{\textbf{L}} \textit{The monk }
\end{dialogue}

\begin{dialogue}
\speak{\textbf{F}}\textit{A white, yellow, and blue bus being towed by a blue tow truck}
\speak{\textbf{L}} \textit{Yellow/white bus being towed by blue}
\end{dialogue}

\section{Reference Chain Extraction}
\label{app:chains}

As explained in Section~\ref{sec:chains}, instead of collecting co-reference chains from manual annotation, we use a heuristics to automatically extract reference chains of dialogue segments likely to contain referring expressions to a chain's target image. 
We consider participants' image labelling actions to signal that a previously discussed target image was identified as either common or different and therefore concluding the current dialogue segment. Due to the spontaneous and unrestricted nature of the PhotoBook dialogues, these labelling actions however do not always indicate segment boundaries as cleanly as possible. To improve the quality of extracted dialogue segments and reference chains, we therefore developed a more context-sensitive heuristics to automate segmentation. The heuristics is implemented as a binary decision tree that uses labelling actions as well as any preceding and subsequent messages and additional labelling actions to better decide on segment boundaries and associated target images. It considers 32 combinations of eight different factors. The first case of the heuristics, for example, states that if
\begin{enumerate}[itemsep=-2pt,leftmargin=14pt] 
    \item the current turn is a message, 
    \item the previous turn was an image labelling action, 
    \item the previous turn was by the other participant, 
    \item the next turn is an image selection action,
    \item the next turn is by the current participant, 
    \item the next labelling action assigns a \textit{common} label, 
    \item the other participant's previous labelling and the current participant's next labelling address the same target image, and
    \item there is a non-empty, currently developing dialogue segment,
\end{enumerate}
then we assume that after one speaker selected an image as common, the other speaker makes one utterance and marks the same image as common, which is resolved by saving the currently developing segment with the common image as referent and initialising a new segment with the trailing utterance of the second speaker. This prevents creating a segment with just the trailing utterance (that cannot be a complete segment) which would be the naive decision if segmenting was based solely on labelling actions. Other cases include whether the next turn is a message by the other participant followed by them labelling a second image as different (likely to indicate that two images were discussed and the segment should be extended by the following message as well as the second target image) or whether none of the preceding and subsequent turns contains labelling actions (indicating an ongoing dialogue segment). 

The following shows a typical example of an automatically extracted chain of dialogue segments associated with the image in Figure \ref{fig:chain_sample}: 

\begin{figure}[t]
    \centering
    \includegraphics[width=0.6\linewidth]{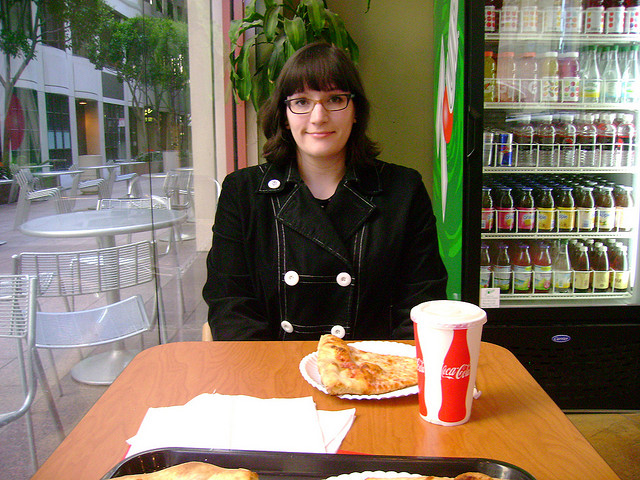}
    \caption{Sample image MS COCO \#449904.}
    \label{fig:chain_sample}
\end{figure}

\begin{dialogue}
\speak{\textbf{B}} \textit{Hello}
\speak{\textbf{A}}  \textit{Hi}
\speak{\textbf{A}} \textit{Do you have a woman with a black coat with buttons, glasses and a piece of pizza on table}
\speak{\textbf{B}} \textit{no} \\
\noindent\makebox[\linewidth]{\rule{\linewidth}{0.4pt}}\\
\speak{\textbf{A}} \textit{Lady with black shirt, glasses with pizza on table?}
\speak{\textbf{B}}  \textit{yes}
\speak{\textbf{A}} \textit{Table with orange bowl with lemons and liquor, cups?}
\speak{\textbf{B}} \textit{no}\\
\noindent\makebox[\linewidth]{\rule{\linewidth}{0.4pt}}
\speak{\textbf{A}} \textit{Orange bowl with lemons, liquor?}
\speak{\textbf{B}} \textit{lady pizza}
\speak{\textbf{A}} \textit{No lady pizza}
\speak{\textbf{B}}  \textit{yes}\\
\noindent\makebox[\linewidth]{\rule{\linewidth}{0.4pt}}
\speak{\textbf{B}}  \textit{woman and pizza} 
\speak{\textbf{A}} \textit{Empty kitchen wood coloured cabinets?} 
\speak{\textbf{A}} \textit{No woman pizza}
\speak{\textbf{B}} \textit{no}
\end{dialogue}

\noindent About 72\% of all segments are assigned to a single co-reference chain, 25\% were automatically assigned to co-reference chains of two different target images and the remaining 3\% to 3 or more chains.

\section{Reference Resolution Experiments}
\label{app:exp}

\paragraph{Data and Results}
In addition to the results reported on Table~\ref{tab:results} in Section~\ref{sec:exp}, which concern the target images in the test set, 
here we report the scores for target images on the validation set (Table~\ref{targetvalmetrics}) and the scores for non-target images (Table~\ref{nontargetmetrics}). 
The latter constitute the large majority of candidate images, and thus results are substantially higher for this class.

\begin{table}[!h]\centering 
\scalebox{0.9}{
\begin{tabular}{|l|c|c|c|}
\hline
\textbf{Model}          & \textbf{Precision} & \textbf{Recall} & \textbf{F1} \\ \hline
\sc No History& 56.37    & 75.91 & 64.70    \\ \hline
\sc History & 56.32    & 78.10  & 65.45   \\ \hline
\sc No image & 34.61    & 62.49  & 44.55   \\ \hline
\end{tabular}}
\caption{Results for target images in the validation set.}
\label{targetvalmetrics}
\end{table}

\begin{table}[t]\centering
\scalebox{0.76}{
\begin{tabular}{|l|c|c|c|}
\hline
\textbf{Model}     & \textbf{Precision} & \textbf{Recall} & \textbf{F1} \\ \hline
\sc No History & 95.34 (95.37)& 89.48 (89.42) & 92.31 (92.30)\\ \hline
\sc History & 95.61 (95.76) & 89.26 (89.10) & 92.33 (92.31)\\ \hline
No image & 92.24 (92.10) & 79.33 (78.74) & 85.30 (84.90)\\ \hline
\end{tabular}}
\caption{Results for non-target images in the test set (and the validation set, in brackets).}
\label{nontargetmetrics}
\end{table}

\pagebreak

\noindent
Finally, Table~\ref{chain_rank} reports the overall number of reference chains in the dataset
broken down by length, that is, by the number of dialogue segments they contain. 

\begin{table}[h]\centering
\scalebox{0.85}{
\begin{tabular}{|l|c|c|c|c|c|c|c|}
\hline
 & \multicolumn{7}{c|}{\bf Length  (\# segments) of the reference chains}\\\hline
\textbf{Split}       & \textbf{1} & \textbf{2} & \textbf{3} & \textbf{4} & \textbf{5} & \textbf{6} & \textbf{7} \\ \hline
Train & 1783       & 1340       & 3400       & 4736       & 1322       & 110        & 3          \\ \hline
Val   & 398        & 295        & 754        & 1057       & 281        & 30         & 1          \\ \hline
Test  & 400        & 296        & 754        & 1057       & 281        & 23         & 0          \\ \hline
\end{tabular}}
\caption{Total number of reference chains per length (i.e., \# segments in the chain) in each of the data splits.}
\label{chain_rank}
\end{table}

\newpage

\begin{figure}[h]
\includegraphics[width=\columnwidth]{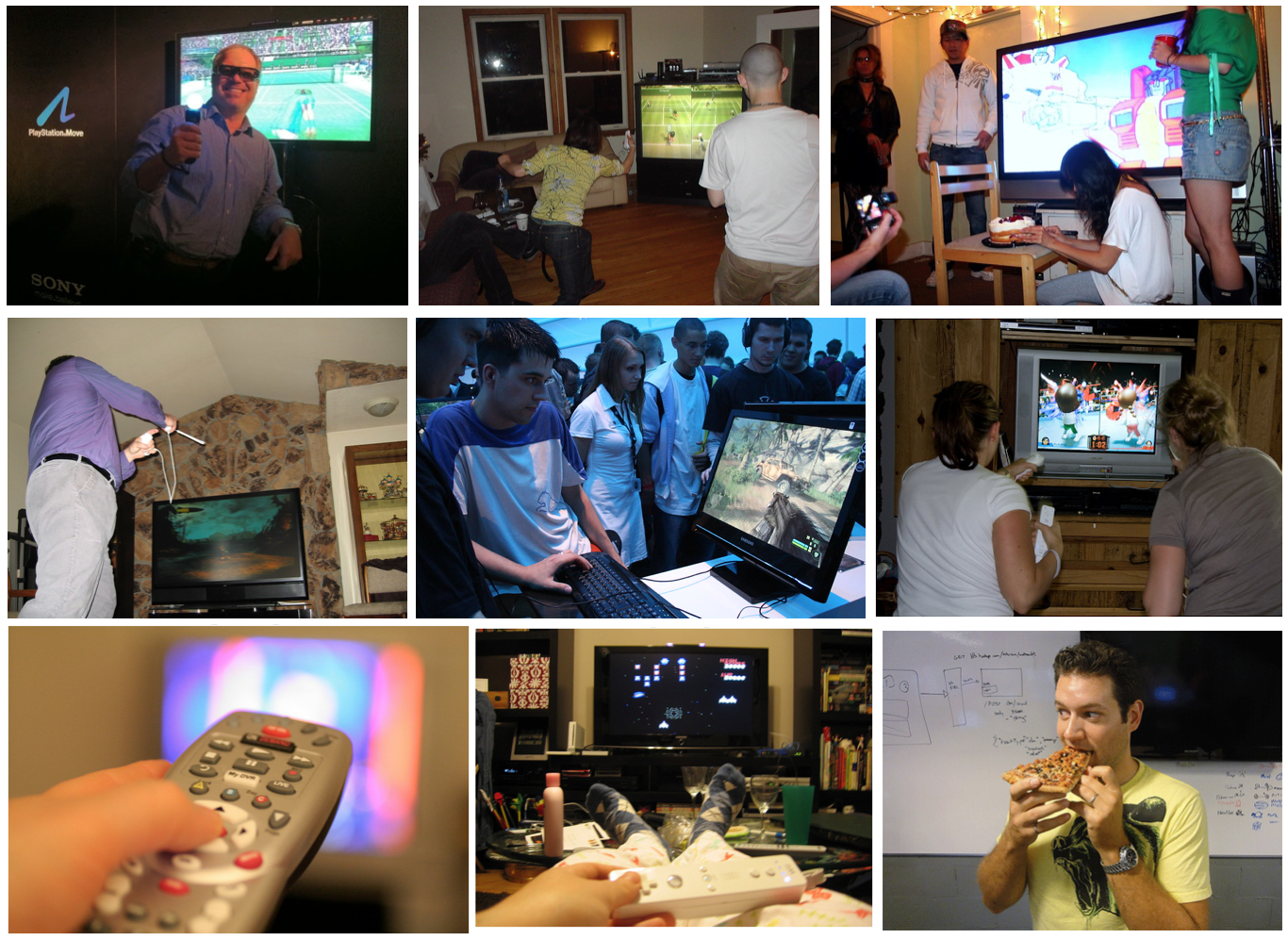} 
\caption{Set of distractors for the target image and segment to be resolved on the left-hand side of Fig.~\ref{fig:ex-history}.
\label{fig:carrot-d}}
\end{figure}

\begin{figure}[h]
\includegraphics[width=\columnwidth]{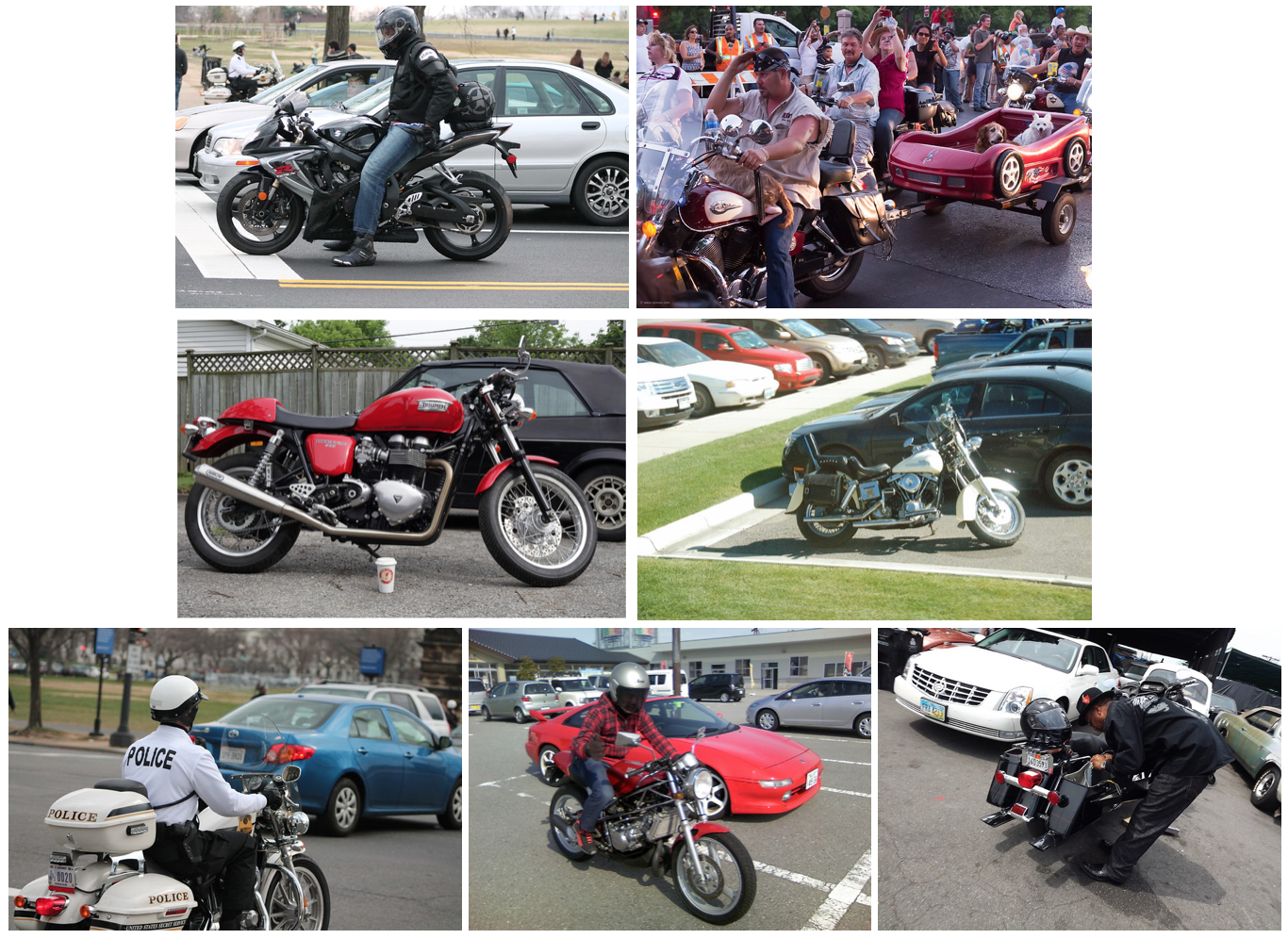} 
\caption{Set of distractors for the target image and segment to be resolved on the right-hand side of Fig.~\ref{fig:ex-history}.
\label{fig:strange-d}}
\end{figure}

\paragraph{Qualitative Analysis} 
Figures~\ref{fig:carrot-d} and~\ref{fig:strange-d} show the distractor images for the examples provided in 
 Figure~\ref{fig:ex-history} and discussed in Section~\ref{sec:qualitative}.

\newpage

\end{document}